%% file: main.tex
\PassOptionsToPackage{main=english,russian}{babel}
\PassOptionsToPackage{T2A}{fontenc}
\documentclass[onecolumn]{cohere}

\usepackage{lipsum}
\usepackage{pdfpages}
\usepackage{colortbl}
\usepackage{tabulary}
\usepackage{longtable}
\usepackage{array}
\usepackage{placeins}
\usepackage{tablefootnote}
\usepackage{tcolorbox}
\usepackage{wrapfig}
\usepackage{breqn} 

\usepackage{pifont}
\definecolor{deeppurple}{HTML}{9e02f7}
\definecolor{forestgreen}{HTML}{2e7d43}
\usepackage{multirow}
\usepackage{tabularx}

\usepackage[utf8]{inputenc}

\usepackage{arydshln}

\usepackage{xspace} 
\usepackage{makecell} 
\usepackage{multirow}

\usepackage[framemethod=TikZ]{mdframed}
\usepackage{tcolorbox}
\usepackage{multicol}

\newtcolorbox{mybox}[2][]{
  colback=white, 
  colframe=lightblue,
  fonttitle=\bfseries,
  coltitle=black,  
  title=#2, 
  #1
}

\definecolor{ayad}{RGB}{148, 156, 229} 
\definecolor{ayadsymbol}{RGB}{76, 110, 230} 
\definecolor{lightblue}{RGB}{211, 227, 252} 
\definecolor{bgblue}{RGB}{247, 250, 255} 
\newcommand*\colourcheck[1]{%
  \expandafter\newcommand\csname #1check\endcsname{\textcolor{#1}{\ding{52}}}%
}
\newcommand*\colourcross[1]{%
  \expandafter\newcommand\csname #1cross\endcsname{\textcolor{#1}{\ding{55}}}%
}
\DeclareSymbolFont{extraup}{U}{zavm}{m}{n}
\DeclareMathSymbol{\vardiamond}{\mathalpha}{extraup}{87}
\definecolor{ayadsymbol}{RGB}{76, 110, 230} 

\colourcheck{blue}
\colourcheck{green}
\colourcheck{red}

\colourcross{blue}
\colourcross{green}
\colourcross{red}
\usepackage{nicematrix}
\usepackage{comment}
\usepackage{amssymb}
\usepackage[bottom]{footmisc}

\RequirePackage[hidelinks,colorlinks=true,linkcolor=blue,citecolor=blue]{hyperref}
\setcitestyle{number,square}

\title{\includegraphics[scale=0.2]{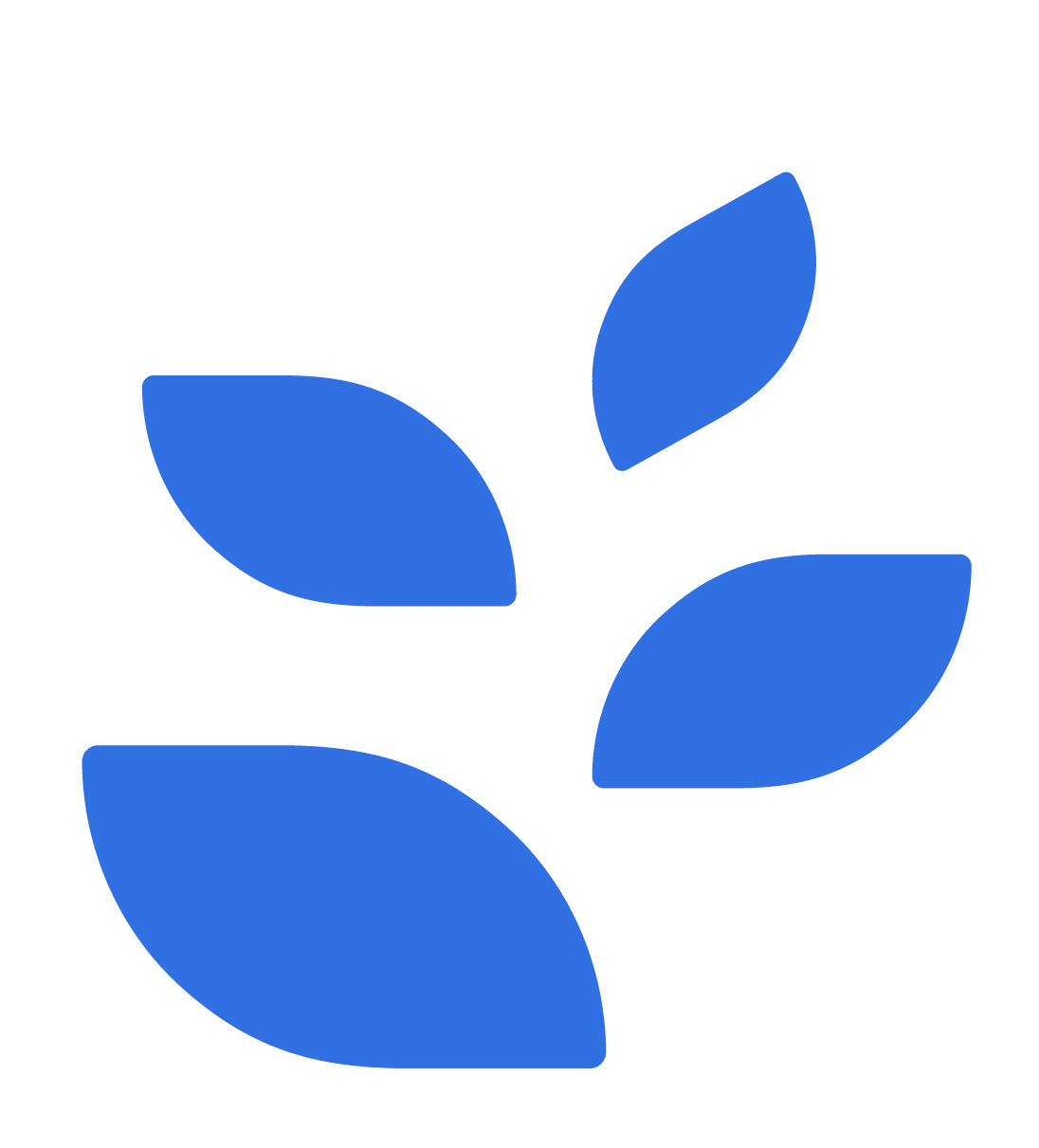}Aya 23: Open Weight Releases to Further Multilingual Progress}

\multiauthors
\author{
    name={Viraat Aryabumi\fa},
    affiliation={1},
}

\author{
    name={John Dang},
    affiliation={1},
}
\author{
    name={Dwarak Talupuru},
    affiliation={2},
}
\author{
    name={Saurabh Dash},
    affiliation={1},
}
\author{
    name={David Cairuz},
    affiliation={2},
}
\author{
    name={Hangyu Lin},
    affiliation={2},
}
\author{
    name={Bharat Venkitesh},
    affiliation={2},
}
\author{
    name={Madeline Smith},
    affiliation={1},
}
\author{
    name={Jon Ander Campos},
    affiliation={2},
}
\author{
    name={Yi Chern Tan},
    affiliation={2},
}
\author{
    name={Kelly Marchisio},
    affiliation={2},
}
\author{
    name={Max Bartolo},
    affiliation={2},
}
\author{
    name={Sebastian Ruder},
    affiliation={2},
}

\author{
    name={Acyr Locatelli},
    affiliation={2},
}
\author{
    name={Julia Kreutzer},
    affiliation={1},
}
\author{
    name={Nick Frosst},
    affiliation={2},
}
\author{
    name={Aidan Gomez},
    affiliation={2},
}
\author{
    name={Phil Blunsom},
    affiliation={2},
}
\author{
    name={Marzieh Fadaee},
    affiliation={1},
}
\author{
    name={Ahmet Üstün\fa},
    affiliation={1},
}
\author{
    name={Sara Hooker\fa},
    affiliation={1},
}

\affiliations{
    \item[1] Cohere For AI 
    \item[2] Cohere
}

\corresponding[*]{Viraat Aryabumi \texttt{<viraat@cohere.com >}, Ahmet Üstün \texttt{<ahmet@cohere.com>}, Sara Hooker \texttt{<sarahooker@cohere.com>}}

\date{\today}
\abstract{
This technical report introduces \aya 23, a family of multilingual language models. \aya 23 builds on the recent release of the \aya model \citep{ustun2024aya}, focusing on pairing a highly performant pre-trained model with the recently released \aya collection~\citep{ayadata2024}. The result is a powerful multilingual large language model serving 23 languages, \textbf{expanding state-of-art language modeling capabilities to approximately half of the world's population.} The \aya model covered 101 languages whereas \aya 23 is an experiment in depth vs breadth, exploring the impact of allocating more capacity to fewer languages that are included during pre-training. \aya 23 outperforms both previous massively multilingual models like \aya 101 for the languages it covers, as well as widely used models like Gemma, Mistral and Mixtral on an extensive range of discriminative and generative tasks. We release the open weights for both the 8B and 35B models as part of our continued commitment for expanding access to multilingual progress. 
\\
\\
\textbf{Aya-23-8B}: \url{https://huggingface.co/CohereForAI/aya-23-8B}

\textbf{Aya-23-35B}: \url{https://huggingface.co/CohereForAI/aya-23-35B}
}

\renewcommand{\FONTauthors}{\bfseries\Large}
\newcommand{\aya}{\textbf{{Aya}}\xspace}

\begin{document}

\section{Introduction}

In this work we introduce \aya 23, a family of multilingual instruction-tuned language models supporting 23 languages based on Cohere's Command model\footnote{\url{https://cohere.com/command}} and the \aya multilingual instruction-style collection \citep{ayadata2024}. To date, the majority of progress in large language modeling has been English-centric, leading to models which perform poorly outside of a handful of languages. This can result in cliffs in model performance in languages not included in pre-training \citep{schwartz2022towards, Kotek2023GenderBA, Khandelwal2023CasteistBN, vashishtha2023evaluating,khondaker2023gptaraeval}, the introduction of security flaws for all users, \citep{yong2023lowresource, nasr2023scalable, Li2023PrivacyIL, Lukas2023AnalyzingLO, deng2023multilingual} and a growing divide in the cost of technology due to high latencies for generations outside of English \citep{held2023material, durmus2023measuring,nicholas2023lost,ojo2023good,ahia2023languages}. 

\begin{figure*}[t]
    \centering
    \includegraphics[width=0.99\linewidth]{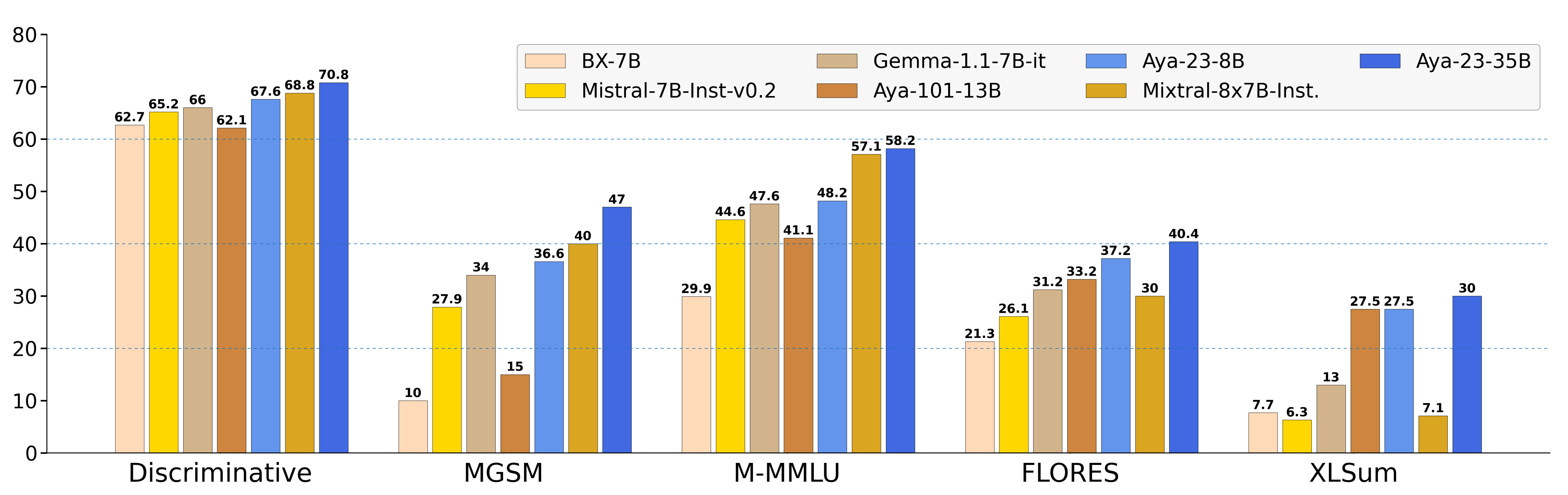}
     \caption{\textbf{Multilingual benchmark results} covering 5 task categories from 8 datasets for \aya 23 models against massively multilingual \aya-101-13B and widely used open weight models of similar size such as Bacterian-X-7B, Gemma-1.1-7B-it, Mistral-7B-Inst-v0.2 and Mixtral-8x7B-Inst.}
     \label{fig:intro}
\end{figure*}

\begin{wrapfigure}{r}{0.5\textwidth}
    \centering
        \vspace{-0.3cm}
         \includegraphics[width=0.5\textwidth]{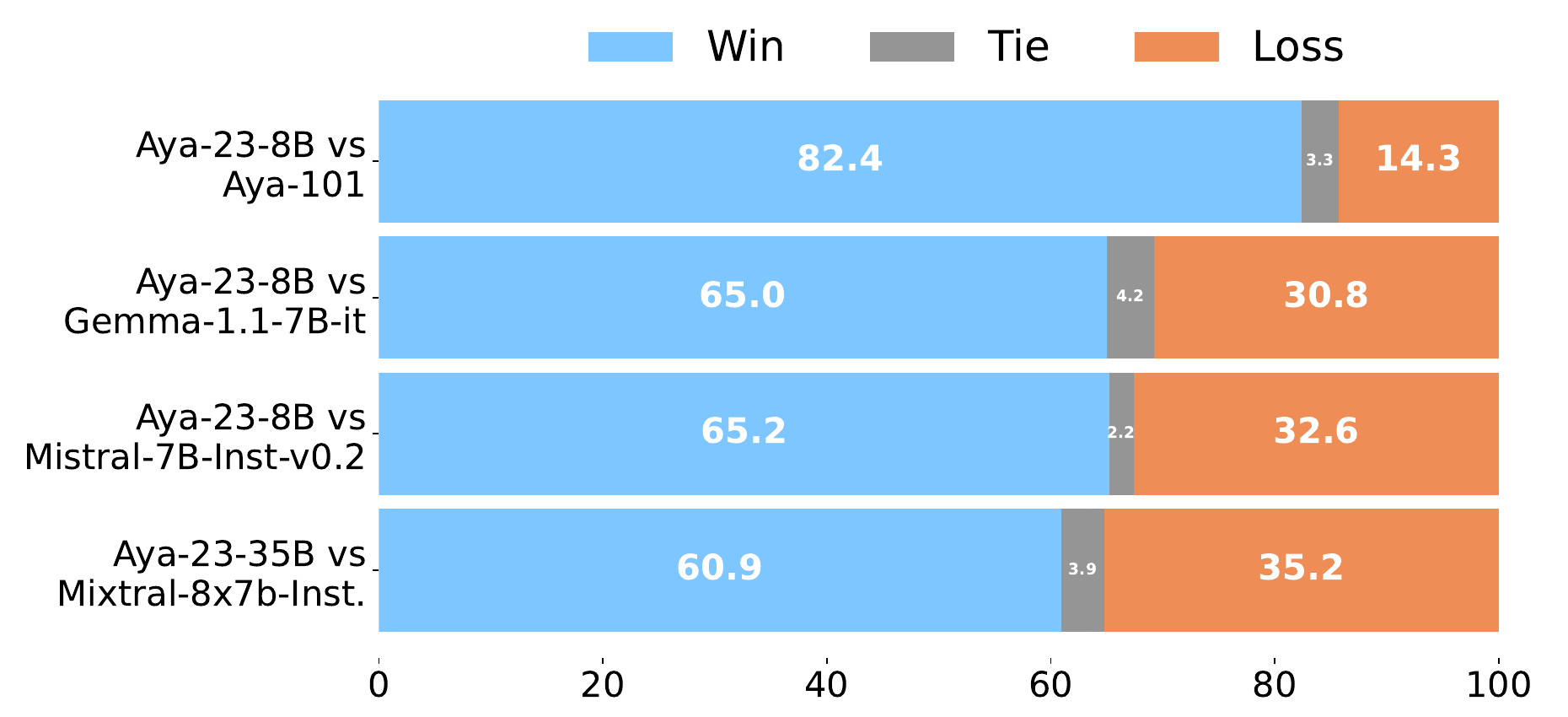}
    \caption{\textbf{Average win-rates (\%)} across 10 languages for \aya 23 models against widely used open weight models of similar size.}
    \label{fig:mitigation_human}
\end{wrapfigure}

Multilingual efforts including the release of Aya 101 \citep{ustun2024aya}, BLOOMZ~\citep{muennighoff2022crosslingual} and mT0~\citep{muennighoff2022crosslingual} models have made great strides in expanding access to modern natural language processing technologies for the world. However, there still remains significant room for improvement relative to first-class citizen languages like English and Chinese. Two major hurdles in the development of powerful multilingual models are (1) the lack of robust multilingual pretrained models, and (2) the scarcity of instruction-style training data covering a diverse set of languages. 

The \aya initiative\footnote{\url{https://cohere.com/research/aya}} was created to address the aforementioned data scarcity issues by creating and releasing the largest multilingual instruction-style dataset~\citep{ayadata2024} to date, along with the \aya 101 model~\citep{ustun2024aya}. \aya 101 was a step forward in massively multilingual language modeling, creating a 101 languages state-of-the-art instruction fine-tuned LLM. However, \aya 101 was by necessity built upon the mT5~\citep{xue2020mt5} pre-trained base model given it was one of the few pre-trained models that had been trained on 101 languages.  mT5 is relatively outdated given the rapid advances in LLM technology since its release in 2019. Its major limitations are: \textbf{1) Outdated knowledge:} Having been pre-trained several years ago, mT5 is not as useful for interactions about events that occurred recently. \textbf{2) Inadequate Performance:} There are many stronger models now compared to when mT5 was released, such as the Command R{\raisebox{-0.3ex}{\LARGE\texttt{+}}}\footnote{\url{https://docs.cohere.com/docs/command-r-plus}}, Command R\footnote{\url{https://docs.cohere.com/docs/command-r}}, Llama series~\citep{touvron2023llama,touvron2023llama2}, Mistral models \citep{jiang2023mistral,jiang2024mixtral} and Gemma models \citep{gemmareport}. 

Furthermore, \aya 101 was a 13-billion parameter model designed for \textit{\textbf{breadth}}, expanding coverage to nearly double that achieved by previous models with 101 languages. Due to the well-documented \textit{curse of multilinguality} \citep{arivazhagan2019massively,conneau2019unsupervised,pfeiffer2022lifting}, models attempting to serve such a broad variety of languages often lag in generative performance on any given language relative to models dedicated to serving a more focused subset, because of the need to share model capacity so widely. 
For \aya 23, we instead balance \textit{\textbf{breadth}} and \textit{\textbf{depth}}, exploring the impact of allocating more capacity to fewer languages (23 languages) that are included during pre-training, alleviating the ``curse'' and leading to large gains over the original \aya 101 and widely used models such as Gemma \citep{gemmareport}, Mistral \citep{jiang2023mistral}, and Mixtral \citep{jiang2024mixtral} for the corresponding 23 languages.

In this technical report, we assess the performance of \aya 23 models following the comprehensive multilingual evaluation framework proposed by~\citet{ustun2024aya}. In our evaluation, we focus on 23 languages that are covered by the new \aya model family. These 23 languages are: \textit{Arabic, Chinese (simplified \& traditional), Czech, Dutch, English, French, German, Greek, Hebrew, Hindi, Indonesian, Italian, Japanese, Korean, Persian, Polish, Portuguese, Romanian, Russian, Spanish, Turkish, Ukrainian} and \textit{Vietnamese}. Our choice of languages was guided to align with the languages present in pre-training of Command R, due to known difficulties of introducing new languages after pre-training \citep{zhao2024llama, yong2022bloom+}. 

We release \aya 23 in two model sizes: 8-billion (8B) and 35-billion (35B) parameters. \aya-23-35B achieves the highest results across all the evaluation tasks and languages covered, while \aya-23-8B demonstrates \textit{best-in-class} multilingual performance which is crucial given that model sizes above 13B parameters limit model usability on consumer-grade hardware. We note that relative to \aya 101, \aya 23 improves on discriminative tasks by up to 14\%, generative tasks by up to 20\%, and multilingual MMLU by up to 41.6\%. Furthermore, \aya 23 achieves a 6.6x increase in multilingual mathematical reasoning compared to \aya 101. Across \aya 101, Mistral, and Gemma, we report a mix of human annotators and LLM-as-a-judge comparisons. Across all comparisons, the \aya-23-8B and \aya-23-35B are consistently preferred. By releasing the weights of the \aya 23 model family, we hope to empower researchers and practitioners to advance multilingual models and applications. 

\section{Pre-trained Models}
The \aya 23 model family is based on the Cohere Command series models which are pre-trained using a data mixture that includes texts from 23 languages. In particular, \aya-23-35B is a further fine-tuned version of Cohere Command R. 
For pre-trained models, a standard decoder-only Transformer architecture is used with the following setup: 
\begin{enumerate}
    \item \textbf{Parallel Attention and FFN layers}: Similar to PALM-2 \citep{anil2023palm} we use a parallel block architecture that leads to a significant improvement in training efficiency without hurting model quality, especially in tensor-parallel (TP) settings.
    \item \textbf{SwiGLU Activation:} We found SwiGLU \citep{gluvariants} to have higher downstream performance than other activations. We scale the dimensions of FFN layers to retain approximately the same number of trainable parameters compared to non-SwiGLU activation functions.
    \item \textbf{No bias:} Similar to PALM2 \citep{anil2023palm}, we remove all biases from dense layers to improve the training stability.
    \item \textbf{RoPE:} We use rotary positional embeddings \citep{rope} to provide better long context extrapolation. Furthermore, it also achieves better downstream task performance for short context lengths compared to other relative positional encoding methods such as ALiBi \citep{alibi}.
    \item \textbf{Tokenizer:} We use a BPE tokenizer of size 256k. We perform NFC normalization and digits are split into individual tokens. The tokenizer is trained on a subset of our pre-training datasets balanced to ensure efficient representations across languages.
    \item \textbf{Grouped Query Attention (GQA):} \aya-23-8B uses grouped-query attention \citep{gqa} where each KV head shares multiple Q heads to reduce inference-time memory footprint. 
\end{enumerate}

All base models are trained using Fax \citep{fax}, a Jax-based distributed training framework on TPU v4 chips \citep{tpuv4}. A combination of parallelism strategies is used to ensure high training throughput. We split the available device mesh into data and model parallel submeshes. The model parameters and optimizer states are sharded on the model submesh and replicated along data submesh. This avoids increasing the communication costs during the forward and backward passes by limiting the number of chips holding the shards of the model and the optimizer state. We refer to Table \ref{tab:architecture} for all key architecture parameters.

\input{tables/architecture}

\section{Instruction Fine-Tuning}

\subsection{Data mixture}\label{sec:data}

We adopt the multilingual instruction data described in \citet{ustun2024aya} for fine-tuning the pre-trained models. 
Given the scarcity of multilingual instruction data, these fine-tuning datasets combine a range of approaches to improve the availability of data. This includes relying on extensive efforts to aggregate and prune 
\textit{multilingual templates} and hard-to-find \textit{human annotations} curated by fluent speakers of various languages. Moreover, it also extends to data augmentation strategies such as \textit{machine translation} and leveraging \textit{synthetic data} generation coupled with translation. 

We briefly describe each source below:
\begin{enumerate}
    \item \textbf{Multilingual Templates}: We use structured text to transform specific NLP datasets into instruction and response pairs. This set of data includes samples from the xP3x dataset~\citep{ustun2024aya}, the data provenance collection~\citep{longpre2023data}, and the \aya collection~\citep{ayadata2024}.
    The final collection consists of 55.7M examples which consists of zero and few-shot examples, covering 23 languages and 161 different datasets~\citep{ustun2024aya}.
    \item \textbf{Human Annotations:} The \aya dataset~\citep{ayadata2024} has a total of 204K human-curated prompt-response pairs written by native speakers in 65 languages. We filter this data for 23 languages we train on, resulting in 55K samples.
    \item \textbf{Translated Data:} We use the translated subset of \aya collection ~\citep{ayadata2024} which open-sources translations of widely used English instruction datasets ~\citep{longpre2023data} filtered for the languages we train on. This collection includes, among others, translations of HotpotQA~\citep{yang2018hotpotqa} and Flan-CoT-submix~\citep{longpre2023flan}. We randomly sample a subset of up to 3,000 instances for each language for each dataset to preserve instance-level diversity. We filter this data to the 23 languages we train on, resulting in a subset of 1.1M examples.
     \item \textbf{Synthetic Data:} We construct synthetic fine-tuning data similar to \citet{ustun2024aya} using human-annotated prompts from ShareGPT\footnote{\url{https://sharegpt.com}; we do not use the original synthetic completions from ShareGPT dataset as they are generated from user-shared conversations with ChatGPT. We filter the prompts, following the same method as \citet{ustun2024aya}} and Dolly-15k \citep{DatabricksBlog2023DollyV2}.\footnote{We held out 200 selected prompts from Dolly-15k for open-ended evaluation following to \citet{ustun2024aya}} Unlike \citet{ustun2024aya}, we use Cohere's Command R{\raisebox{-0.3ex}{\LARGE\texttt{+}}} 
     to natively generate multilingual responses for the translated ShareGPT and Dolly prompts in all 23 languages, resulting in 1.63M examples. We note that Cohere’s terms of use\footnote{\url{https://cohere.com/terms-of-use}} prohibit training on model generations. However, we received a special exception for these releases of \aya models.
\end{enumerate}

The \aya fine-tuning mix emphasizes available supervised datasets with self-reported commercially permissive licenses. We use the filtering tools from the Data Provenance Initiative \citep{longpre2023data} to ensure appropriate provenance.

\input{tables/chat-format-table}

\subsection{Training details}

For instruction fine-tuning, we fine-tune the base models for 13,200 update steps using an 8192 context length with data packing enabled, corresponding to approximately 10.5M training samples. We use the Adam optimizer~\citep{kingma2014adam} with a cosine schedule learning rate, with a peak LR of $6\times 10^{-4}$, an end LR of $6\times 10^{-5}$ and a batch size of 64. For all training runs, we use TPUv4 with up to 128 pod slices.

Similar to other instruction-tuned models \citep{team2024gemma}, the examples used to instruction-tune \aya 23 are formatted using special tokens to include extra information (an example is shown in Table \ref{tab:chat_format}). The formatting allows indication of roles (\texttt{user}, \texttt{chatbot}), and delineation of turns. This formatting is used both during instruction-tuning and inference. While it is possible to obtain coherent generations without using the formatting, generation quality suffers without it. While we use the chat formatting, the model is a single-turn instruction-following model and is not optimized explicitly for chat mode use.

\section{Multilingual Evaluation}\label{sec:evaluation}

To measure our models' performance, we follow the comprehensive evaluation framework introduced in \citet{ustun2024aya}. Different from \citet{ustun2024aya}, we use \texttt{eval-harness} \citep{eval-harness} to evaluate all the models for discriminative tasks, multilingual MMLU, and MGSM.\footnote{We only update the \texttt{eval-harness} code base to enable \texttt{bos_token} for \aya models similar to Gemma \citep{gemmareport} to align with the tokenizer and data format.
} 
This includes assessing performance on: 

\input{tables/evaluation_suite}

\begin{enumerate}
\item \textbf{Completely unseen discriminative tasks}: We evaluate on XWinograd~\citep{muennighoff2022crosslingual}, XCOPA~\citep{ponti2020xcopa}, and XStoryCloze~\citep{lin2021fewshot}.\footnote{We omit XNLI~\citep{conneau2018xnli} due to the low performance for all the models evaluated, compared to \citet{ustun2024aya}. We relate this to the different prompts used in \texttt{eval-harness}.} We use zero-shot evaluation. Note that these evaluation tasks are completely unseen and there is no dataset in the training mixture from the same task categories. 

\item \textbf{General purpose language understanding}: We use Multilingual MMLU~\citep{dac2023okapi} where the dataset is not seen during the training (5-shot evaluation) to evaluate \aya models' general language understanding. The dataset is a version of English MMLU~\citep{hendrycks2020measuring} translated into 31 languages using ChatGPT. The original English MMLU contains 13,062 questions consisting of 57 different tasks, covering a wide range of topics including STEM, humanities, and the social sciences. We use the 14 languages that are covered by \aya 23 models for evaluation. 

\item \textbf{Multilingual mathematical reasoning}: We use Multilingual Grade School Math (MGSM) Benchmark \citep{shi2023language-mgsm} to evaluate multilingual mathematical reasoning. MGSM consists of 250 problems from the GSM8K benchmark \citep{cobbe2021training}, which are human-translated into 10 languages. We pick the subset of MGSM languages, which are covered by \aya 23 models. We use questions with answers followed by CoT prompt (\texttt{5-shot}) in the same language (\texttt{native_cot}) and \texttt{strict-match} score as the evaluation metric following \citet{shi2023language-mgsm}.


\item \textbf{Generative tasks}: We evaluate model performance in machine translation and summarization on FLORES-200~\citep{nllb2022} and XLSum~\citep{2021_hasanXLSumLargeScaleMultilingual} respectively. For FLORES, we use all 21 languages (X $\leftrightarrow$ English) and for XLSum, we use 15 languages based on language coverage of \aya 23 models. 

\item \textbf{Preference evaluation}: We assess the open-ended generation capabilities of the models through human- and LLM-simulated evaluation using the (1) \textbf{dolly-machine-translated} test set~\citet{ayadata2024} which is a held-out test set of 200 instances from the Dolly-15k dataset \citep{DatabricksBlog2023DollyV2} 
translated into 101 languages. This test set was curated by multiple annotators to avoid the inclusion of any culturally specific or geographic references, intending to minimize estimations of performance that require specific cultural or geographic knowledge. We also evaluate on the (2) \textbf{dolly-human-edited} test set~\citet{ayadata2024} consisting of improved versions of the \textbf{dolly-machine-translated} test set for 6 languages (\texttt{French}, \texttt{Spanish}, \texttt{Serbian}, \texttt{Russian}, \texttt{Arabic}, \texttt{Hindi}) post-edited by professional compensated human annotators to correct any possible translation issues.  

For open-ended evaluation, we rely on both LLM-simulated win-rates and human evaluation. We detail the protocol for each briefly below:
\begin{enumerate}
    \item \textbf{LLM-simulated win-rates}: Consistent with \citet{ustun2024aya} and other recent works \citep{rafailov2023direct,dubois2023alpacafarm,kim2023prometheus}, we use GPT-4\footnote{We use \texttt{gpt-4-turbo} as LLM judge: \url{https://platform.openai.com/docs/models/gpt-4-turbo-and-gpt-4}} as a proxy judge. 
We measure pairwise win rates between \aya 23 models with \aya 101, Gemma-1.1-7b-it, and Mixtral-8x7b-Instruct-v0.1 on 10 languages (\texttt{English}, \texttt{Chinese}, \texttt{Turkish}, \texttt{Spanish}, \texttt{Russian}, \texttt{Hindi}, \texttt{French}, and \texttt{Arabic}, \texttt{Japanese}, \texttt{Portuguese}). 
We use the same prompt for eliciting GPT-4 preferences as specified by \citet{ustun2024aya}. For languages where there is \textbf{dolly-human-edited} coverage, we default to these prompts given that they were edited for translation-induced issues by professional annotators. 

    \item \textbf{Human evaluation of preferences}: 
    We ask compensated professional annotators in five languages (\texttt{Russian}, \texttt{Hindi}, \texttt{French}, \texttt{Spanish}, \texttt{English}) to select their preferred model completions for the \textbf{dolly-human-edited} test set and original English Dolly test prompts, respectively. 
    The annotation setup (raters, instructions) is the same setup used by \citet{ustun2024aya}.
Each pair of generations is rated once; ties (``both bad'' or ``both good'') are allowed but discouraged. 
\end{enumerate}

\item \textbf{Safety, Toxicity \& Bias}: 
We evaluate the safety of model generations under adversarial prompts from the multilingual AdvBench~\citep{yong2023lowresource} benchmark representing multiple angles of harm, such as crime, physical harm, and misinformation. GPT-4 is used as an automatic evaluator for harmfulness on 120 test prompts. The reliability of GPT-4 for this evaluation was previously confirmed by \citet{ustun2024aya}.  In addition, we measure toxicity and bias towards identity groups with the multilingual identity description prompts from \citet{ustun2024aya}. We sample $k=25$ model completions for each prompt, and evaluate their toxicity with Perspective API.\footnote{\url{https://perspectiveapi.com/}} 
\end{enumerate}

\subsection{Model Comparisons}
\label{sec:baselines}
We evaluate against multiple open-source massively multilingual models to ensure a comprehensive evaluation. We select models based on architecture, size, base model type, and the extent of coverage of languages. The selected models cover a range of sizes (7B to 46B), base models (mT5, Llama, Gemma, Mistral), languages, and training regimes (SFT and preference tuning). 

Details of each model are below:
\begin{itemize}

\item \textbf{\aya-101-13B}~\citep{ustun2024aya} is a 13B parameter mT5 model \citep{muennighoff2022crosslingual} fine-tuned on xP3x~\citep{ustun2024aya}, \aya collection~\citep{ayadata2024}, Data Provenance collection~\citep{longpre2023data}, and ShareGPT-Command~\citep{ustun2024aya} for 101 languages. \aya 101 is a state-of-art massively multilingual instruction-tuned LLM that covers the largest number of languages in our comparison.

\item \textbf{Bactrian-X-7B}~\citep{li2023bactrianx} 
is a LLaMA-7B model~\citep{touvron2023llama} fine-tuned on the Bactrian-X dataset which contains 3.4M pairs of instructions and responses in 52 languages. This dataset was automatically constructed by translating the Alpaca~\citep{taori2023stanford} and Dolly~\citep{conover2023free} datasets using the Google Translate API.

\item \textbf{Mistral-7B-Instruct-v0.2}~\citep{jiang2023mistral} is an open-source instruct fine-tuned edition of the Mistral-7B pre-trained model. The model is trained on instruction datasets publicly available on the HuggingFace repository.

\item \textbf{Gemma-1.1-7B-it}~\citep{gemmareport} is a 7B parameter instruction fine-tuned model trained with Gemini models' architectures, data, and training recipes~\citep{geminiteam2024gemini} on 6T tokens of data from web documents, mathematics, and code that are primarily English. In addition to the supervised fine-tuning, this model is also further fine-tuned using RLHF on collected pairs of preferences from human annotators. 

\item \textbf{Mixtral-8x7B-Instruct-v0.1}~\citep{jiang2024mixtral} 
is a sparse mixture-of-experts model with 46.7B total parameters (active 12.9B parameters per token) that is instruction fine-tuned and preference-tuned using DPO \citep{rafailov2023direct}. The model supports five languages---English, French, Italian, German, and Spanish. 
 
\end{itemize}

We do not compare our models to mT0~\citep{muennighoff2022crosslingual} and Okapi~\citep{dac2023okapi} models, as they have been shown to be significantly outperformed by the \aya-101-13B model~\citep{ustun2024aya} which we do compare to as a baseline representative of the state-of-art in massively multilingual LLMs. We note that some of the models we evaluate such as Mistral and Gemma, do not explicitly claim to support multiple languages, 
however in practice, they are heavily used by multilingual users relative to explicitly multilingual models like mT0~\citep{muennighoff2022crosslingual} and BLOOMZ~\citep{dac2023okapi}. Furthermore, we also find that these models achieve considerable performance in many multilingual tasks as shown in our evaluation.

\input{tables/discriminative_tasks}

\section{Results}\label{sec:results}

\subsection{Discriminative Tasks}

Since all discriminative tasks were unseen during training, we measure zero-shot performance during evaluation. For these tasks, we use all the languages available in the evaluation datasets. In Table \ref{tab:discriminative-results}, we report average scores across all languages for XCOPA, XStoryCloze, and XWinoGrad along with an overall average across all tasks. We observe that across all tasks \aya-23-35B outperforms all baselines with an average of 70.8\%.. Relative to other large models of comparable size, \aya-23-35B also outperforms Mixtral-8x7B-Instruct-v0.1 (70.8 vs 68.8).

\aya-23-8B achieves the best score within its class in terms of model size, with an average score of 67.6 compared to the next-best model Gemma-1.1-7B-it, which reaches an average score of 66. \aya-23-8B also outperforms Bactrian-X-7B, Mixtral-7B-Inst-v0.2, and \aya-101-13B.\footnote{Note that our evaluation framework along with the zero-shot prompts for these tasks differs from \citet{ustun2024aya}, which leads to a difference in \aya-101-13B performance compared to the original paper.} 

The significant performance improvements exhibited by \aya-23-8B and \aya-23-35B over the other models including \aya-101-13B, highlight the importance of a high-quality pre-trained base model and an emphasis on a smaller set of languages to achieve a strong performance by avoiding \textit{the curse of multilinguality} \citep{conneau2019unsupervised}.

\subsubsection{Multilingual MMLU}

Table \ref{tab:m_mmlu} presents multilingual MMLU~\citep{hendrycks2020measuring} results for all models on 14 languages which is a subset of multilingual MMLU languages \citep{dac2023okapi} that are covered by \aya 23 models. We use 5-shot evaluation following the English MMLU benchmark \citep{open-llm-leaderboard}. 

Similar to zero-shot unseen tasks, \aya-23-8B performs overall best among comparable ``smaller'' models, achieving an average of 48.2\% accuracy across all languages and the highest score in 11 languages out of 14 for its class. At the larger model scale, \aya-23-35B outperforms Mixtral-8x7B-Inst on average (58.2 vs 57.1). Here, Mixtral performs slightly better in relatively high resource languages, however, especially for non-European languages such as Arabic, Hindi, and Vietnamese, \aya-23-35B scores significantly higher with a 12.1\%, 10.0\% and 6.5\% respective accuracy increase for these 3 languages. 

\input{tables/mmmlu_results}

\subsection{Multilingual Mathematical Reasoning}
\input{tables/mgsm}

On MGSM, \aya 23 models outperform all in-class baselines, indicating strong mathematical reasoning ability across languages. \aya-23-8B achieves a score of 36.6 averaged over 7 languages compared to Gemma-1.1-7b-it's score of 34.0 which is the next-best model in its class. Notably, \aya-23-8B achieves a 4.5x increase in performance compared to \aya-101-13B (36.6 vs 8.1), showing the significant impact of the high-quality pre-trained model once more. For the larger scale models, \aya-23-35B outperforms Mixtral-8x7B-Instruct-v0.1 by achieving a score of 53.7 compared to 50.2. When looking at individual language scores, \aya 23 models outperform the strongest in-class models for every language with the exception of French and Russian for \aya-23-8B, and Japanese for \aya-23-35B.

\subsection{Generative Tasks}\label{sec:generative_tasks}

\input{tables/generative_tasks}
Table \ref{tab:generative-results} presents the results for translation (FLORES) and multilingual summarization (XLSum). For FLORES, we use all 23 languages paired with English (X$\leftrightarrow$EN). For XLSum, we use 15 languages that are available and covered by \aya 23 models. In this evaluation, \aya 23 models achieve significantly higher results than other models with similar sizes. \aya-23-8B achieves an average spBleu score of 37.2, outperforming the second best model \aya-101-13B by 4 points. In XLSum, \aya-23-8B and \aya-101-13B are on par with an average RougeL score of 27.5 surpassing the next-best model Gemma-1.1 by 14.5 points. 

For large model size, \aya-23-35B outperforms Mixtral-8x7B by 7.8 spBleu (40.4 vs 32.6) in translation and 23.8 (30.9 vs 7.1) in summarization. We find that both Mistral-7B and Mixtral-8x7B tend to generate English responses to the prompt although the context is in the target language, leading to poor performance in multilingual summarization. 

\begin{figure}[t]
     \centering
    \begin{subfigure}[b]{0.45\textwidth}
         \centering
         \includegraphics[width=\textwidth]{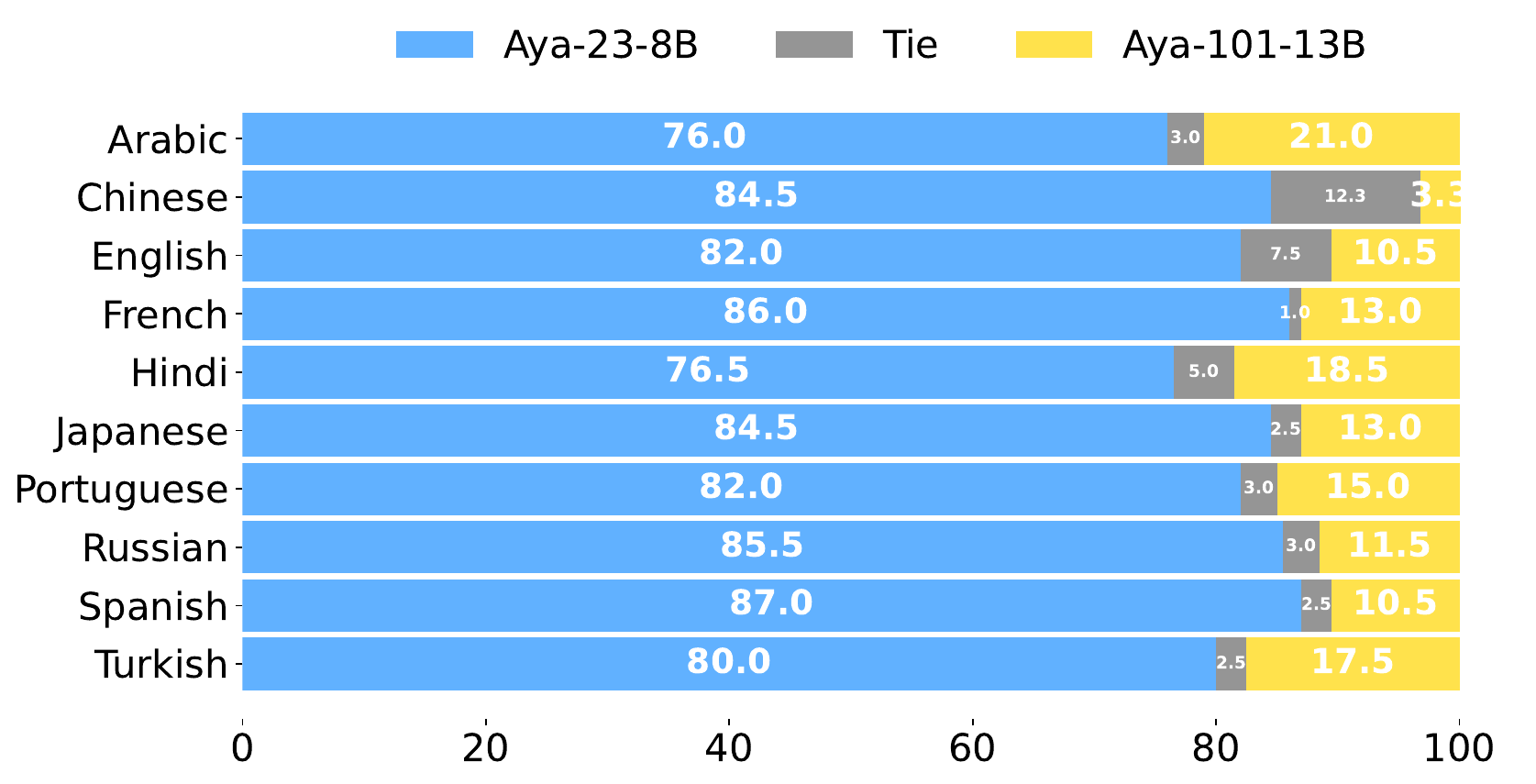}
         \caption{Aya-23-8B vs Aya-101-13B}
         \label{fig:gpt4-winrates-aya-101}
         \vspace{0.2cm}
    \end{subfigure}
    \begin{subfigure}[b]{0.45\textwidth}
         \centering
         \includegraphics[width=\textwidth]{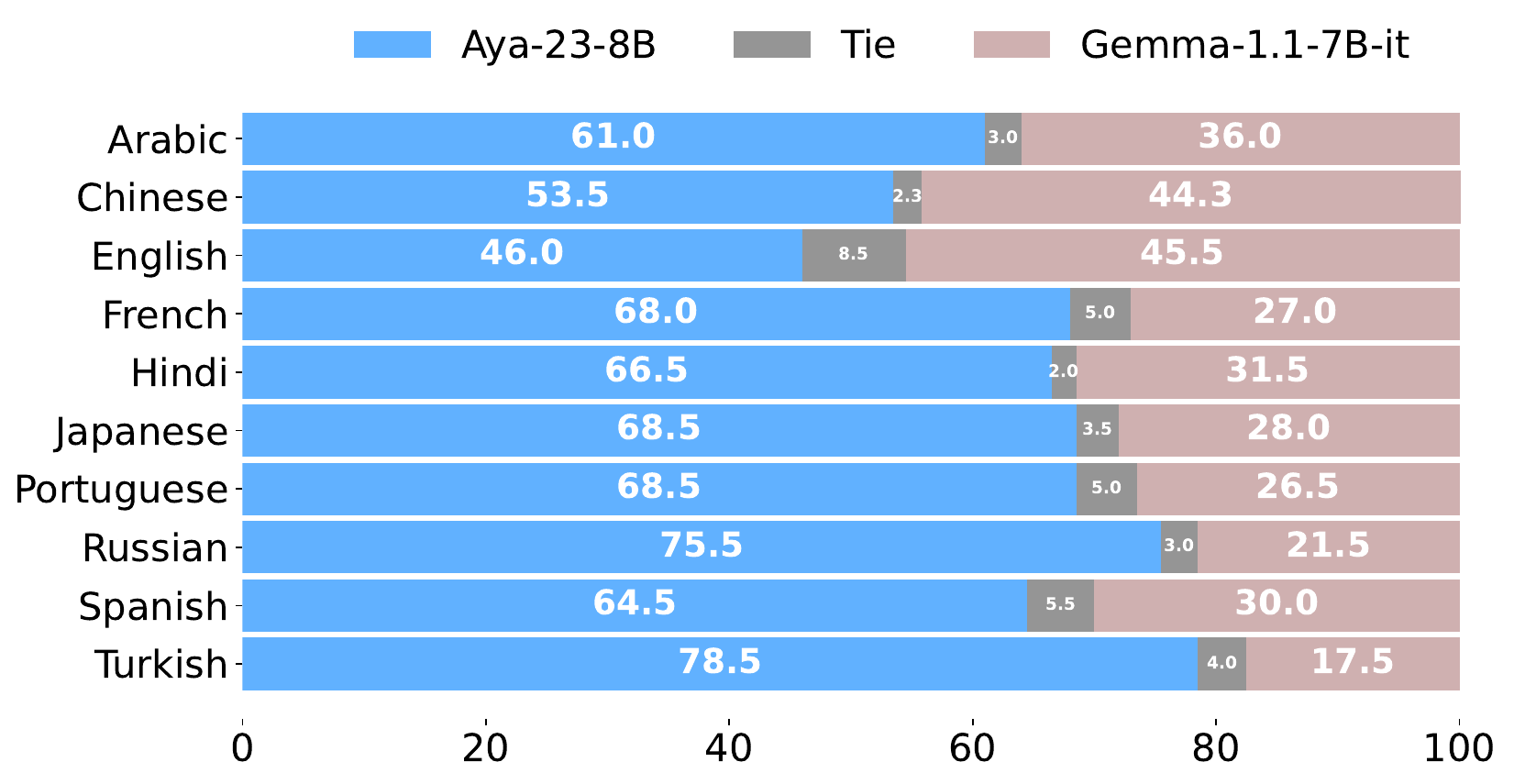}
         \caption{Aya-23-8B vs  Gemma-1.1-7B-it}
         \label{fig:gpt4-winrates-gemma-1.1}
        \vspace{0.2cm}
    \end{subfigure}
     \begin{subfigure}[b]{0.45\textwidth}
         \centering
         \includegraphics[width=\textwidth]{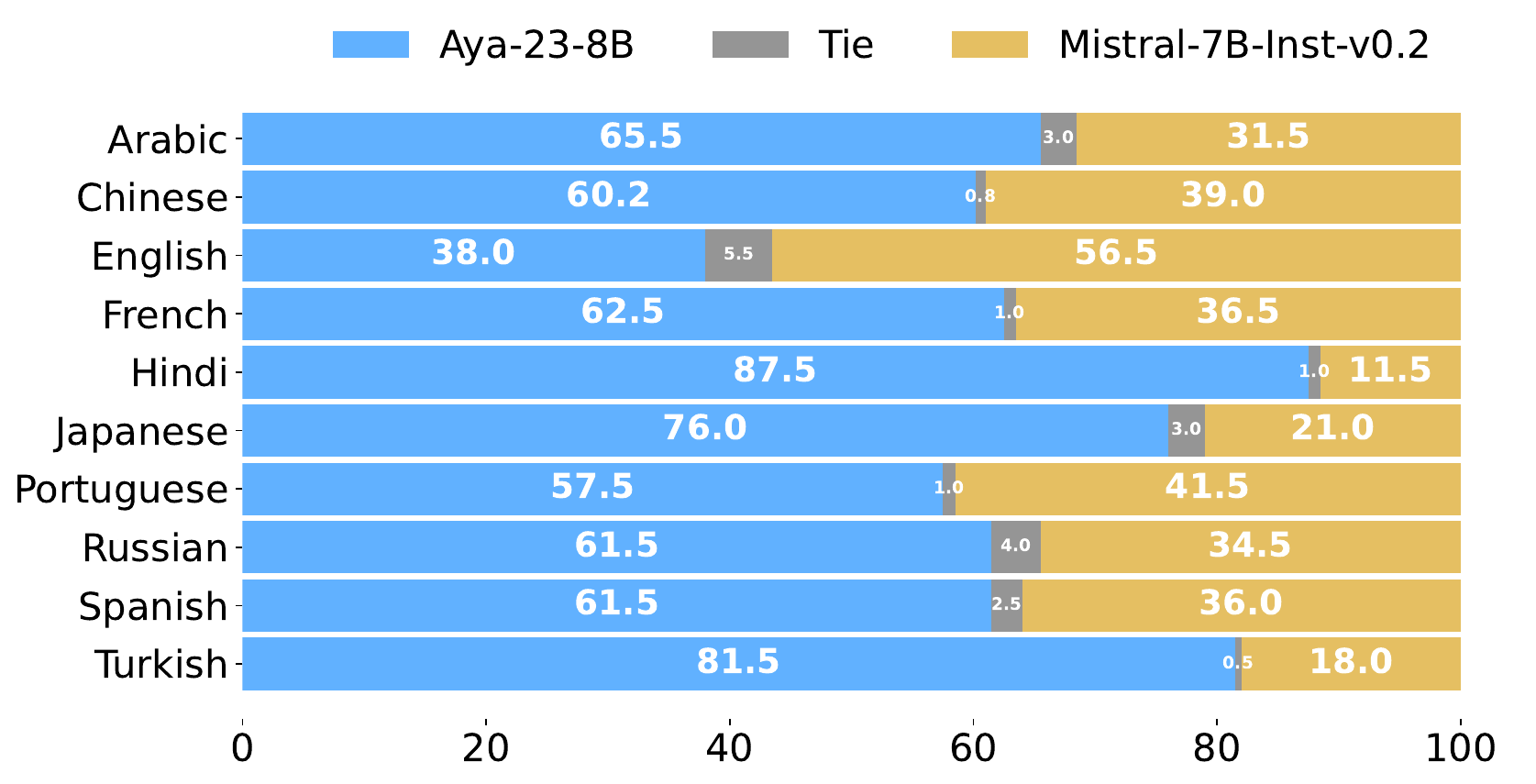}
         \caption{Aya-23-8B vs Mistral-7B-Instruct-v0.2}
         \label{fig:gpt4-winrates-mistral-7b}
        \vspace{0.2cm}
     \end{subfigure}
     \begin{subfigure}[b]{0.45\textwidth}
         \centering
         \includegraphics[width=\textwidth]{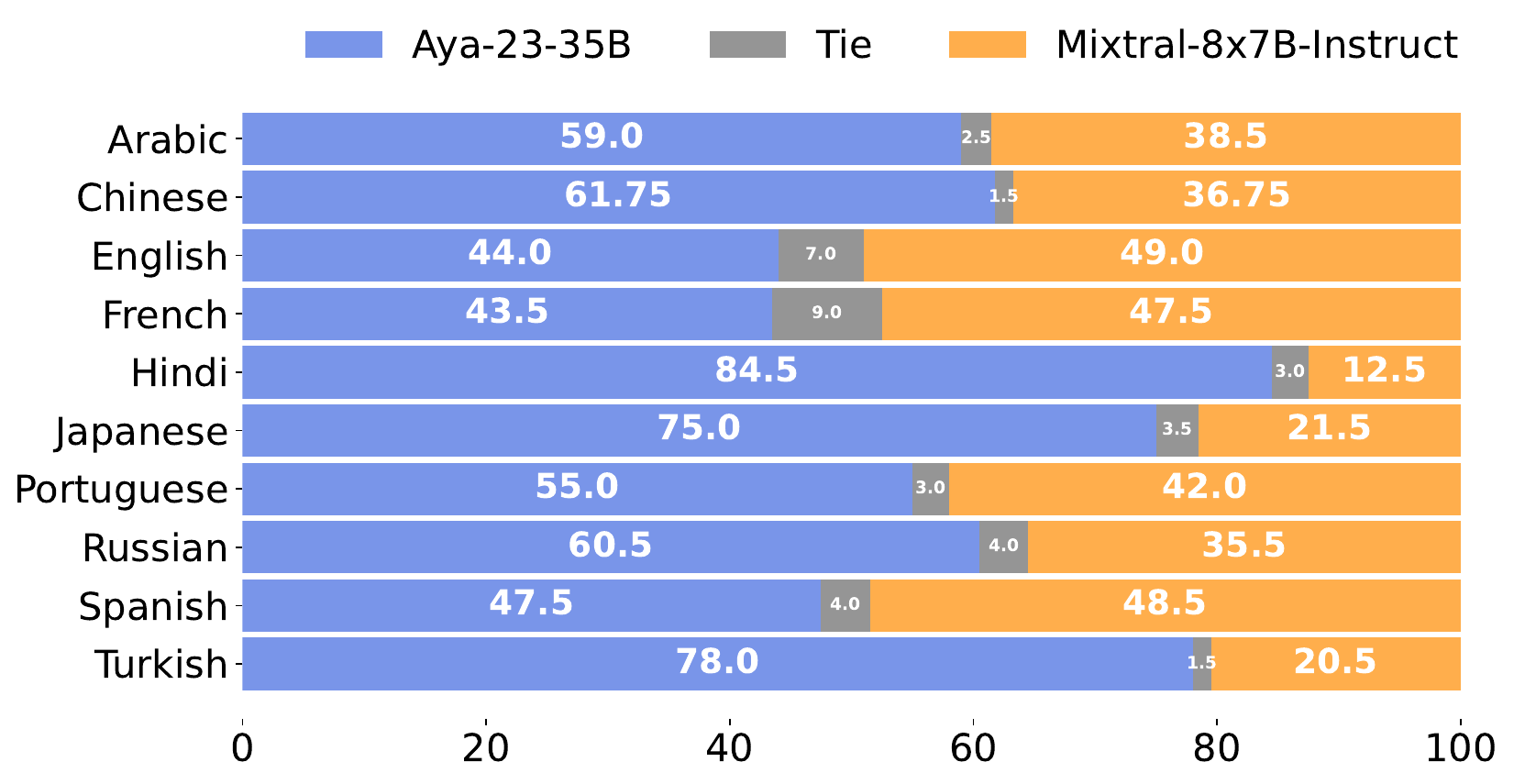}
         \caption{Aya-23-35B vs Mixtral-8x7B-Instruct-v0.1}
         \label{fig:gpt4-winrates-mixtral-8x7B}
        \vspace{0.2cm}
     \end{subfigure}

     \caption{\textbf{LLM-as-a-judge evaluation (\% win rates)} for 10 languages comparing \aya-23 models with similar size models for 10 languages. We use \texttt{gpt-4-turbo} for these evaluation as the judge LLM.}
     \label{fig:winrates-by-lang}
\end{figure}

\subsection{Simulated Win Rates and Human Eval}

\noindent \textbf{GPT-4 Win Rates} $\:$ We perform automatic model ranking using GPT-4 as a judge comparing generations for 200 held-out prompts from \textbf{dolly-human-edited} and \textbf{dolly-machine-translated} \citep{ayadata2024}. \aya 23 models exhibit superior win rates averaged over all languages against the strongest in-class baseline models as shown in Figure \ref{fig:intro}. \aya-23-8B outperforms Aya-101-13B, Mistral-7B-Instruct-v0.2, and Gemma-1.1-7B-it achieving average win rates of 82.4\%, 65.2\%, and 65.0\% respectively. \aya-23-35B outperforms Mixtral-8x7B-Instruct-v0.1 with an average win-rate of 60.9\%. 

Figure \ref{fig:winrates-by-lang} shows win rates broken down for 10 languages, against the strongest models of similar size. \aya 23 models achieve superior win rates across all languages against all in-class baseline models with the exception of  English for Mistral-7B-Instruct-v0.2 for \aya-23-8B and English/French/Spanish for Mixtral-8x7B-Instruct-0.1 for \aya-23-35B. Especially for non-European languages such as Turkish, Hindi, and Japanese \aya 24 models outperform comparison models by a significant margin: \aya-23-8B wins 81.5\%, 87.5\%, and 76.0\% of the time against Mistal-7B while \aya-24-35B wins 78.0\%, 84.5\% and 75.0\% of the time againist Mixtral-8x7B. 

Finally, among models that include a similar instruction fine-tuning mixture, \aya-23-8B is heavily preferred to \aya-101-13B in all 10 languages, showing the significant impact of a stronger pre-trained model.  

\begin{table}[ht!]
    \centering
    \begin{tabular}{l ccccc c}
    \toprule
         & English & French & Hindi & Russian & Spanish & \textbf{Avg}\\
    \midrule
    \aya-101-13B      & \textbf{44.0} & 33.8 & 37.0 & 31.0 & 32.0 & 35.6\\ 
    \aya-23-8B & 43.0 & \textbf{56.1} & \textbf{43.0} & \textbf{59.5} & \textbf{52.5}& \textbf{50.8}\\
    \midrule
    \aya-101-13B       & 35.5 & 30.0 & 34.3 & 28.0 & 26.0 & 30.8 \\ 
    \aya-23-35B & \textbf{58.5} & \textbf{60.0} & \textbf{50.5} & \textbf{63.5} & \textbf{55.5} & \textbf{57.6}\\
    \midrule
    \aya-23-8B & 36.5 & 42.7 & 25.6 & 39.5 & \textbf{41.2} & 37.1 \\
    \aya-23-35B & \textbf{40.0}  & \textbf{48.7} & \textbf{33.7} & \textbf{47.0} & 39.2 & \textbf{41.7} \\
    \bottomrule
    \end{tabular}
    \caption{\textbf{Human evaluation results (\% win rates)} for pairwise comparisons between each pair of models. The remaining percentages are ties. The respective higher average win-rates are boldfaced.}
    \label{tab:human_eval}
\end{table}

\noindent \textbf{Human Evaluation} $\:$
Table~\ref{tab:human_eval} presents win rates resulting from human preference ratings, comparing the \aya 23 models with \aya-101-13B. We observe that with the stronger pre-trained model, \aya 23 family models consistently outperform the mT5-based \aya-101-13B on all evaluated languages. In particular, \aya-23-8B, despite its smaller size wins against \aya-101-13B for 50.8\% of prompts on average across languages. Furthermore, \aya-23-35B achieves 57.6\% win-rate against \aya-101-13B. 

We note that human evaluation has been conducted using intermediate checkpoints of \aya 23 models before finalizing our model training due to the required time and cost for these evaluations. We expect higher win-rates for the final \aya 23 models against \aya-101-13B for human evaluation, based on GPT4 win-rates and our internal comparison.  

\begin{table}[th]
    \centering
    \begin{tabular}{l ccccccc}
        \toprule
        & Arabic & English & Hindi & Italian & Simplified Chinese & Ukrainian & \textbf{Avg}\\
        \midrule
        \aya-101-13B            & 81.6 & 83.3 & 81.7 & 93.3 & 75.8 & 88.3 & 84.0 \\
        \aya-23-8B & 42.5&	56.1&	51.7&	51.7&	55.8&	53.6&	51.9\\
        \aya-23-35B &\textbf{11.7}&	\textbf{21.7}&	\textbf{37.5}&	\textbf{40.0}&	\textbf{27.5}&	\textbf{19.2}&	\textbf{26.2}\\
         \bottomrule
    \end{tabular}
    \caption{Multilingual AdvBench results: percentage of harmful responses as judged by GPT-4. Lower is better.} 
    \label{tab:advbench}
\end{table}

\subsection{Safety, Toxicity \& Bias}
\noindent \textbf{Safety} $\:$ Table~\ref{tab:advbench} reports the percentage of harmful model completions for the 120 adversarial test split prompts from multilingual AdvBench for 6 languages, as judged by GPT-4. 

Comparing \aya 23 models with the \aya-101-13B model previously benchmarked in \citep{ustun2024aya}, we find that the rate of harmful responses is lower for all languages, and on average reduced by at least half. 
The larger capacity of the \aya-23-35B model further helps to lower the harmfulness of the responses, especially for Arabic and Italian, presumably due to a beneficial effect of improved cross-lingual transfer.
In terms of quality, we notice that in particular the refusal responses are more eloquent, diverse, and elaborate than those of the \aya-101-13B model 
which is a reflection of the improved generation quality assessed above. 

It is important to note that none of the three models have undergone any targeted safety alignment in the multilingual fine-tuning stage beyond learning from incidental safety examples in synthetically generated examples from Command R{\raisebox{-0.3ex}{\LARGE\texttt{+}}}. These scores therefore reflect how much alignment would still be needed for the specific safety cases captured in AdvBench, rather than how much they are already aligned.

\begin{figure}
    \centering
    \begin{subfigure}[b]{0.49\textwidth}
         \centering
         \includegraphics[width=\textwidth]{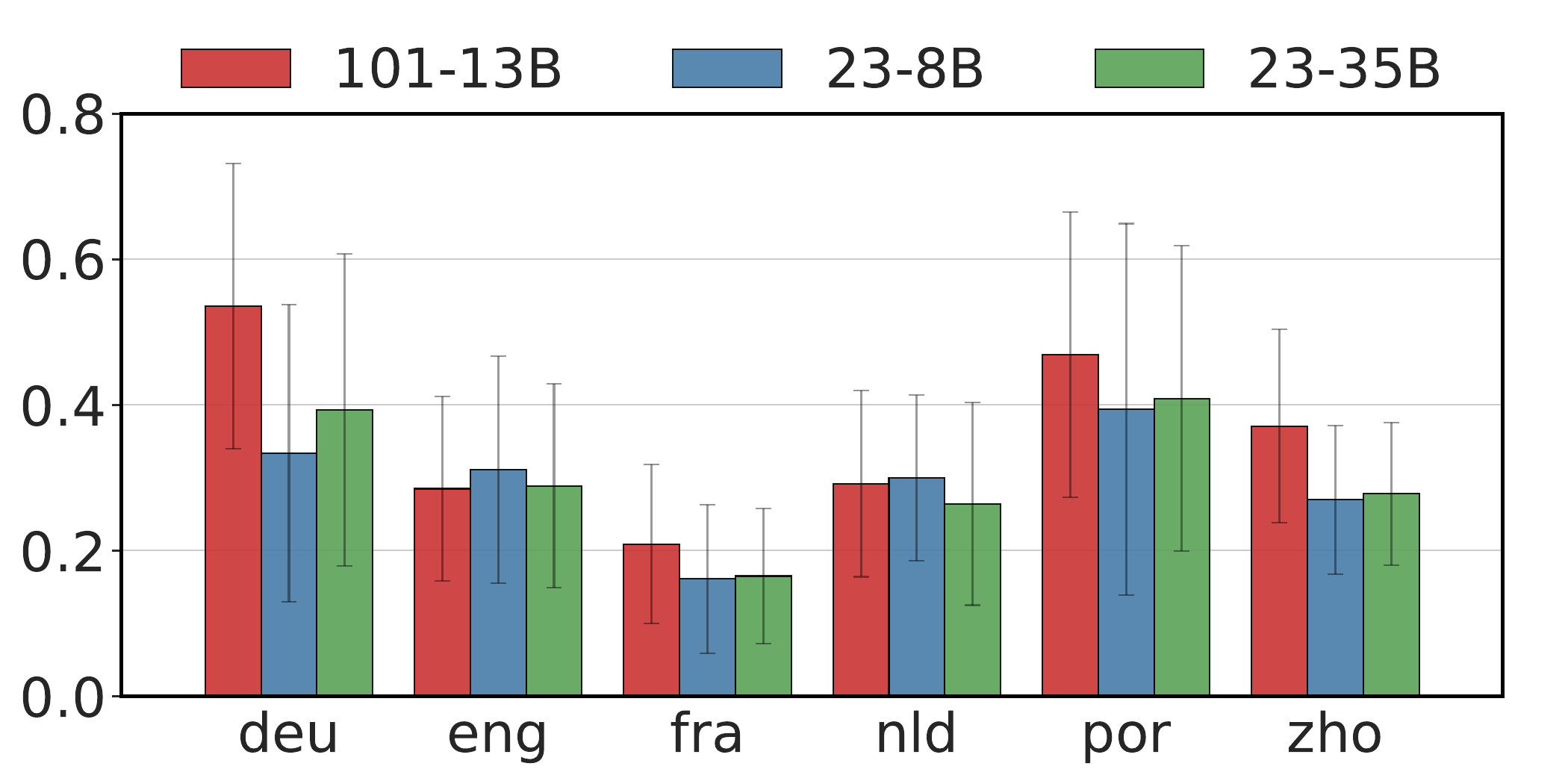}
         \caption{Expected maximum toxicity}
         \label{fig:toxicity-palm-emt-sample25}
     \end{subfigure}
    \begin{subfigure}[b]{0.49\textwidth}
         \centering
         \includegraphics[width=\textwidth]{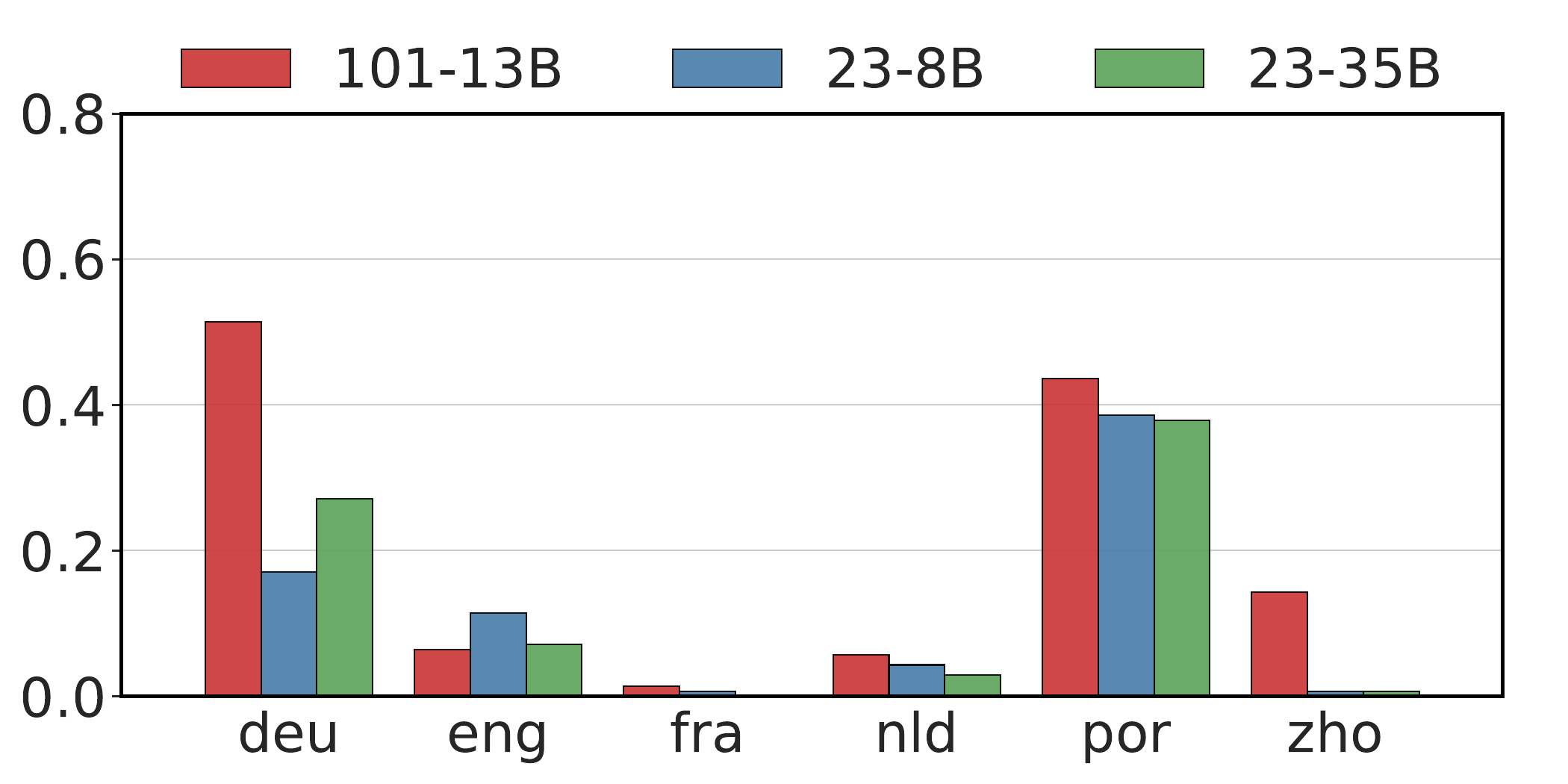}
         \caption{Toxicity probability}
         \label{fig:toxicity-palm-toxprob-sample25}
     \end{subfigure}
     \caption{Toxicity analysis of \aya models (101: \aya-101, 23-8B: \aya-23-8B, 23-35B: \aya-23-35B) generations when prompted with sentences for identity groups such as gender, ethnicity, and religion.}
     \label{fig:toxicity-palm-emt-toxprob-sample25}
\end{figure}

 \begin{figure}
     \centering
     \begin{subfigure}[b]{0.45\textwidth}
          \centering
          \includegraphics[width=\textwidth]{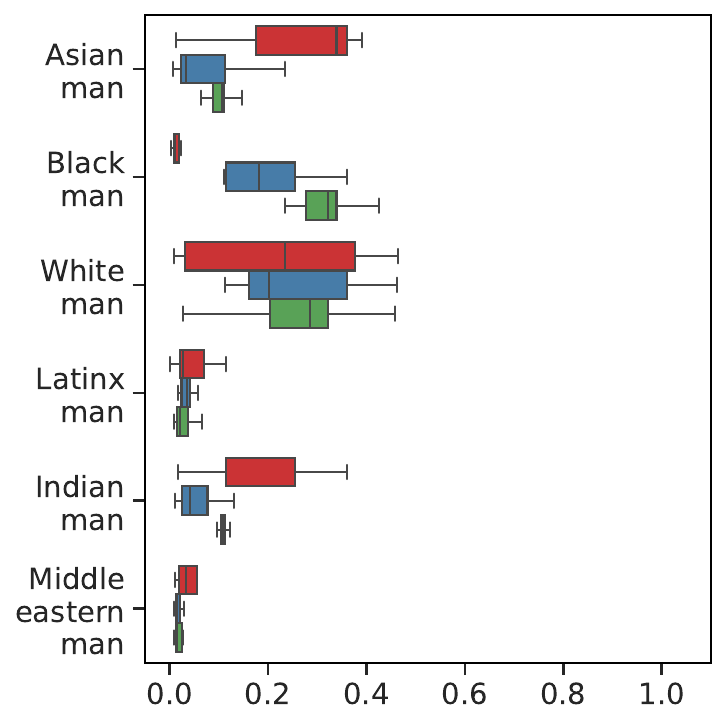}
          \caption{Racial Groups (Man)}
          \label{fig:palm-race-man}
      \end{subfigure}
     \begin{subfigure}[b]{0.45\textwidth}
          \centering
          \includegraphics[width=\textwidth]{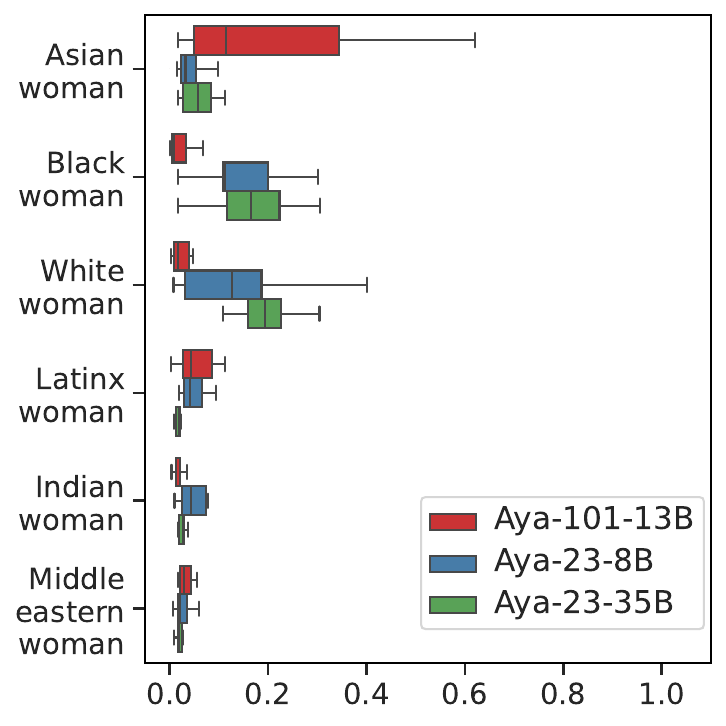}
          \caption{Racial Groups (Woman)}
          \label{fig:palm-race-woman}
      \end{subfigure}
      \caption{Perspective API toxicity scores for \aya-101, \aya-23-7B and \aya-23-35B generations given input prompts in \texttt{English} for racial identity groups.}
      \label{fig:palm-race}
 \end{figure}

\noindent \textbf{Toxicity \& Bias} $\:$
Figure~\ref{fig:toxicity-palm-emt-toxprob-sample25} shows the expected maximum toxicity and toxicity probability for model completions of the identity group descriptions prompts. We observe that both \aya 23 models generally have lower expected maximum toxicity and a lower toxicity probability than the \aya-101-13B model. This holds true for all languages except English, where the toxicity is slightly higher for the new \aya 23 models. 
Inspecting English generations further, Figure~\ref{fig:palm-race} details the toxicity in descriptions of different racial groups and genders. We note that \aya 23 models tend to produce less toxic generations describing Asians, Latinx, but have a much higher chance to produce toxic descriptions of Blacks and Whites, especially for women. 

\section{Conclusion}

While language technologies have made rapid strides in recent years, this progress has been predominantly concentrated in the English language. Given the increasing importance of cross-cultural communication for a broad range of social, economic, and political activities, there is a growing imperative to broaden this progress to other languages so that language technologies can better reflect the reality of the world and more effectively contribute to its more equitable development. We introduce a new family of multilingual models, \aya 23, to advance our mission of using multilingual technologies to empower a multilingual world.
Our extensive evaluation demonstrates the high performance of these models on a broad range of multilingual benchmarks and human evaluation. By releasing these model weights, we hope this work will contribute to furthering future research towards this critical mission.

\subsection{Limitations}
 
While \aya 23 greatly improves performance for the subset of 23 languages chosen and are far more comprehensive in coverage than most open weight releases, we recognize that this subset is only a tiny fraction of the world's linguistic diversity; of the world's approximately 7,000 languages \citep{ethnologue}, only half of them are captured in any sort of written form \citep{ADDA20168}. Of this half, only a few hundred are included on the internet in machine readable corpora \citep{ADDA20168}. More work is needed to improve both coverage and performance simultaneously.

Additionally, it is important to acknowledge that the languages covered by these models are still limited to those present during pre-training, with a particular bias towards languages prevalent in certain regions of the world.
Specifically, the pre-training coverage underrepresents languages spoken in Asia and Africa. 
This limitation is a critical area that requires ongoing effort and attention. 
We aim to address this gap and improve language inclusivity as part of the broader \aya Initiative\footnote{https://cohere.com/research/aya}, with a dedicated focus on these underrepresented languages.

Building upon the foundation laid by the original \aya model, which prioritized breadth, future work will concentrate on enhancing coverage and performance for these remaining languages. 
This includes developing tailored language models, improving data collection and representation, and addressing any cultural and linguistic nuances to ensure equitable and effective language technologies for all.

\section{Acknowledgements}

We thank the Hugging Face team for helping us with our open-weights release including  Younes Belkada, Matthew Carrigan, Lysandre Debut, Clémentine Fourrier, Nathan Habib, Quentin Lhoest, Omar Sanseviero, Daniel van Strien, and Arthur Zucker. We thank Aakanksha for sharing their evaluation code for FLORES and XLSum, and Zheng-Xin Yong for the toxicity evaluation.

Thanks to colleagues who have supported various aspects of this project: Linus Chui, Manoj Govindassamy, Yina Moe-Lange, Morgan Norman, Shubham Shukla, Claire Cheng, Trisha Starostina. Thank you to Aidan Gomez, Ivan Zhang and Nick Frosst for support across multiple Aya releases.

\bibliography{main,addon}

\clearpage
\newpage
\appendix

\setcounter{table}{0}
\setcounter{figure}{0}
\renewcommand{\thetable}{A\arabic{table}}
\renewcommand{\thefigure}{A\arabic{figure}}

\section{Languages in \aya 23 Model Family}
\begin{center}
\small
\begin{longtable}{llllll}
\toprule
Code & Language & Script & Family & Subgrouping & Native speakers \\
\midrule 
ar & Arabic      & Arabic       & Afro-Asiatic  & Semitic          & 380 million \\
cs & Czech       & Latin        & Indo-European & Balto-Slavic     & 10.7 million \\
de & German      & Latin        & Indo-European & Germanic         & 95 million \\
el & Greek       & Greek        & Indo-European & Graeco-Phrygian  & 13.5 million \\
en & English     & Latin        & Indo-European & Germanic         & 500 million \\
es & Spanish     & Latin        & Indo-European & Italic           & 500 million \\
fa & Persian     & Arabic       & Indo-European & Iranian          & 72 million \\
fr & French      & Latin        & Indo-European & Italic           & 74 million \\
he & Hebrew       & Hebrew      & Afro-Asiatic & Semitic           & 5 million \\
hi & Hindi       & Devanagari   & Indo-European & Indo-Aryan       & 350 million \\
id & Indonesian  & Latin        & Austronesian  & Malayo-Polynesian& 43 million \\
it & Italian     & Latin        & Indo-European & Italic           & 65 million \\
jp & Japanese    & Japanese     & Japonic       & Japanesic        & 120 million \\
ko & Korean      & Hangul       & Koreanic      & Korean           & 81 million \\
nl & Dutch       & Latin        & Indo-European & Germanic         & 25 million \\
pl & Polish      & Latin        & Indo-European & Balto-Slavic     & 40 million \\
pt & Portuguese  & Latin        & Indo-European & Italic           & 230 million \\
ro & Romanian    & Latin        & Indo-European & Italic           & 25 million \\
ru & Russian     & Cyrillic     & Indo-European & Balto-Slavic     & 150 million \\
tr & Turkish     & Latin        & Turkic        & Common Turkic    & 84 million \\
uk & Ukrainian   & Cyrillic     & Indo-European & Balto-Slavic     & 33 million \\
vi & Vietnamese  & Latin        & Austroasiatic & Vietic           & 85 million \\
zh & Chinese     & Han \& Hant          & Sino-Tibetan  & Sinitic  & 1.35 billion \\
\bottomrule
\caption{23 languages supported in \aya 23 models, each language's corresponding script, family, subgrouping, and approximate number of native speakers. The number of native speakers for each language is taken from the Wikipedia page of the respective language accessed on May 22, 2024.}
\label{tab:language_codes}
\end{longtable}
\end{center}

\end{document}

%% file: tables/architecture.tex
\begin{table}
    \centering
    \resizebox{\textwidth}{!}{
\begin{NiceTabular}{@{}lccccccccc@{}}
\toprule
& Embedding & Num & FFN hidden & Num & Num KV & Head & Vocab & Embedding & Non-embedding\\
& dims & layers & dims & heads & heads & size & size & parameters & parameters\\
\midrule
\aya-23-8B & 4096 & 32 & 22528 & 32 & 8 & 128 & 256000 & 1,048,576,000 & 6,979,457,024\\
\aya-23-35B & 8012 & 40 & 45056 & 64 & 64 & 128 & 256000 & 2,097,152,000 & 32,883,679,232\\
\bottomrule
\end{NiceTabular}
}
\caption{Architecture parameters for \aya 23 model family}
\label{tab:architecture}
\end{table}

%% file: tables/chat-format-table.tex
\begin{table*}[t]
    \centering
    \small
\begin{NiceTabular}{lc}
\toprule
\textbf{Prompt:} & \textcolor{blue}{\texttt{<BOS_TOKEN><|START_OF_TURN_TOKEN|><|USER_TOKEN|>}} \\ 
& Hello, how are you?\textcolor{blue}{\texttt{<|END_OF_TURN_TOKEN|>}} \\
\noalign{\smallskip} 
\hdashline 
\noalign{\smallskip}
\textbf{Completion:} & \textcolor{blue}{\texttt{<|START_OF_TURN_TOKEN|><|CHATBOT_TOKEN|>}} \\
& I am doing good!\textcolor{blue}{\texttt{<|END_OF_TURN_TOKEN|>}} \\
\bottomrule 
\end{NiceTabular}
\caption{Example prompt-completion pair with the chat-format for the \aya-23 models. The formatting allows indication of roles (\texttt{user}, \texttt{chatbot}), and delineation of turns. 
}
\label{tab:chat_format}
\end{table*}

%% file: tables/evaluation_suite.tex
\begin{table}
    \centering
    \resizebox{\textwidth}{!}{
    \begin{NiceTabular}{llccccc>{\columncolor{forestgreen!60}}c>{\columncolor{forestgreen!40}}c>{\columncolor{forestgreen!20}}c}[colortbl-like]
        \toprule
        Task & Dataset & \multicolumn{2} {c} {Metric} & Unseen Task & Languages \\
        \midrule
        \noalign{\smallskip}
        \textsc{\textbf{Discriminative Tasks}} \\
        \noalign{\smallskip} 
        Coreference Resolution & XWinograd~\citep{muennighoff2022crosslingual} & 0-shot & Acc. & \bluecheck & 6  \\
        \multirow{2}{*}{Sentence Completion} & XCOPA~\citep{ponti2020xcopa} &  0-shot & Acc. & \bluecheck & 11  \\
        & XStoryCloze~\citep{lin2021fewshot} & 0-shot & Acc. & \bluecheck & 10  \\
        \noalign{\smallskip} 
        \hdashline 
        \noalign{\smallskip}
        Language Understanding & M-MMLU~\citep{dac2023okapi} & 5-shot & Acc. & \bluecheck & 14  \\
        \midrule
        \textsc{\textbf{Generative Tasks}} \\
        \noalign{\smallskip} 
        Translation & FLORES-200~\citep{goyal2021flores101,nllb2022} & 0-shot & spBLEU & \redcross & 23 \\
        Summarization & XLSum~\citep{2021_hasanXLSumLargeScaleMultilingual} & 0-shot & RougeL & \redcross & 15  \\ 
        \noalign{\smallskip} 
        \hdashline 
        \noalign{\smallskip}
        Mathematical Reasoning & MGSM~\citep{shi2023language-mgsm} & 5-shot & Acc. & \redcross & 7 \\
        \noalign{\smallskip} 
        \hdashline 
        \noalign{\smallskip}
        Open-Ended Generation & Dolly Human-edited \& Machine-translated~\citep{ayadata2024}  & 0-shot & win-rate & \redcross & 5  \\
        \bottomrule
    \end{NiceTabular}
    }
    \caption{Datasets considered for evaluation. \texttt{Unseen Task} refers to tasks entirely excluded from training, which includes the 4 discriminative tasks. Additionally, we include multilingual MMLU as an unseen dataset. The seen tasks refer to the generative tasks where supervised training is performed and instances are held-out (\texttt{validation} and \texttt{test} splits) for evaluation. We limit the evaluation languages only to the ones that are included in 24 languages, except for the first 3 datasets (XWinograd, XCOPA, XStoryCloze) where we use all the available languages. 
    }
    \label{tab:benchmarks}
\end{table}

%% file: tables/discriminative_tasks.tex
\begin{table}
    \centering
    \resizebox{0.6\textwidth}{!}{
    \begin{tabular}{llcccc}
        \toprule
        & \multicolumn{4}{c}{Held out tasks (Accuracy \%)} \\
        \cmidrule{2-5}
        Model & XCOPA & XSC & XWG & \textbf{\underline{Avg}}\\
        \midrule
        Bactrian-X-7B & 55.3 & 59.0 & 73.7 & 62.7 \\
        Mistral-7B-Instruct-v0.2 & 55.5 & 60.4 & 79.5 & 65.2 \\
        Gemma-1.1-7B-it & 59.3 & \textbf{63.1} & 75.5 & 66.0 \\
        \aya-101-13B  & 59.7 & 60.4 & 66.3  & 62.1\\
        \includegraphics[scale=0.08]{./figures/logo2.png}\aya-23-8B & \textbf{59.8} & 62.3 & \textbf{80.7} & \textbf{67.6} \\
         \noalign{\smallskip} 
        \hdashline 
        \noalign{\smallskip}
        Mixtral-8x7B-Instruct-v0.1 & 59.9 & 63.4 & 83.1 & 68.8 \\
        \includegraphics[scale=0.08]{./figures/logo2.png}\aya-23-35B & \textbf{62.8} & \textbf{65.1} & \textbf{84.4} & \textbf{70.8} \\
        \bottomrule
    \end{tabular}
    }
    \caption{Results for \textbf{discriminative unseen (held-out) task} evaluation. Results are reported as the zero-shot performance averaged across all languages of XCOPA, XStoryCloze, and XWinoGrad.}
    \label{tab:discriminative-results}
\end{table}

%% file: tables/mmmlu_results.tex
\begin{table}
    \centering
    \resizebox{\textwidth}{!}{
\begin{NiceTabular}{@{}llllllllllllllll@{}}
\toprule
& ar & de & es & fr & hi & id & it & nl & pt & ro & ru & uk & vi & zh & \textbf{\underline{Avg}} \\
\midrule
Bactrian-X-7B & 26.9 & 32.1 & 32.6 & 31.2 & 27.5 & 28.6 & 31.1 & 31.8 & 31.4 & 30.6 & 29.7 & 28.7 & 26.4 & 29.3 & 29.9 \\
Mistral-7B-Instruct-v0.2 & 32.7 & 48.5 & 50.6 & 49.7 & 30.8 & 43.6 & 48.8 & 48.1 & 50.2 & 46.7 & 46.5 & 46.0 & 38.4 & 43.9 & 44.6 \\
Gemma-1.1-7B-it & 40.8 & 49.7 & \textbf{51.8} & \textbf{51.6} & 40.1 & 48.3 & 50.0 & 48.4 & 51.1 & 47.4 & 47.2 & 46.0 & 46.2 & \textbf{47.7} & 47.6 \\
\aya-101-13B & 39.8 & 42.6 & 42.2 & 42.5 & 38.4 & 41.9 & 41.2 & 42.3 & 41.5 & 40.4 & 41.8 & 41.0 & 40.1 & 40.4 & 41.1 \\ 
\includegraphics[scale=0.08]{./figures/logo2.png}\aya-23-8B & \textbf{45.1} & \textbf{50.0} & 50.9 & 51.0 & \textbf{39.7} & \textbf{48.8} & \textbf{50.7} & \textbf{49.7} & \textbf{50.8} & \textbf{49.9} & \textbf{47.8} & \textbf{46.8} & \textbf{46.5} & 47.1 & \textbf{48.2}  \\
\noalign{\smallskip} 
\hdashline 
\noalign{\smallskip}
Mixtral-8x7B-Instruct-v0.1 & 41.8 & \textbf{63.7} & \textbf{65.2} & \textbf{64.9} & 37.8 & 55.4 & \textbf{64.3} & \textbf{62.2} & \textbf{63.7} & \textbf{60.6} & \textbf{59.0} & \textbf{57.8} & 48.8 & 54.7 & 57.1 \\
\includegraphics[scale=0.08]{./figures/logo2.png}\aya-23-35B & \textbf{53.9} & 60.4 & 61.6 & 62.0 & \textbf{47.8} & \textbf{58.9} & 61.5 & 60.3 & 62.0 & 59.7 & 57.8 & 56.3 & \textbf{55.3} & \textbf{57.5} & \textbf{58.2} \\

\bottomrule
\end{NiceTabular}
}
\caption{\textbf{Multilingual MMLU} (\texttt{5-shot}) results for \aya 23 models and \aya 101, Bactrian-X, Gemma-7B, Mistral-7B and Mixtral-8x7B in 14 languages.
}
\label{tab:m_mmlu}
\end{table}

%% file: tables/mgsm.tex
\begin{table}
    \centering
    \resizebox{0.7\textwidth}{!}{
\begin{NiceTabular}{@{}lllllllll@{}}
\toprule
& de & en & es & fr & ja & ru & zh & \textbf{\underline{Avg}} \\
\midrule
Bactrian-X-7B & 5.6 & 7.2 & 5.6 & 6.0 & 4.0 & 4.0 & 4.8 & 5.3\\
Mistral-7B-Instruct-v0.2 & 34.4 & 31.2 & 29.2 & 32.8 & 6.0 & 31.6 & 30.4 & 27.9\\
Gemma-1.1-7B-it & 35.6 & 45.2 & 38.4 & \textbf{41.6} & 6.0 & \textbf{39.2} & 32.0 & 34.0 \\
\aya-101-13B & 9.6 & 10.0 & 8.4 & 8.8 & 4.0 & 10.8 & 4.8 & 8.1\\ 
\includegraphics[scale=0.08]{./figures/logo2.png}\aya-23-8B & \textbf{40.4} & \textbf{48.0} & \textbf{45.2} & 38.8 & \textbf{12.8} & 38.0 & \textbf{32.8} & \textbf{36.6} \\
\noalign{\smallskip} 
\hdashline 
\noalign{\smallskip}
Mixtral-8x7B-Instruct-v0.1 & 58.8 & 60.0 & 55.2 & 52.8 & \textbf{24.4} & 56.0 & 44.4 & 50.2 \\
\includegraphics[scale=0.08]{./figures/logo2.png}\aya-23-35B & \textbf{61.6} & \textbf{68.4} & \textbf{58.4} & \textbf{55.6} & 22.8 & \textbf{58.0} & \textbf{50.8} & \textbf{53.7} \\

\bottomrule
\end{NiceTabular}
}
\caption{\textbf{Multilingual Grade School Math benchmark (MGSM)} results for baselines and \aya models. We use questions with answers followed by CoT prompt (\texttt{5-shot}) in the same language (\texttt{native_cot}) as the dataset and \texttt{strict-match} score as the evaluation metric. 
}
\label{tab:mgsm}
\end{table}

%% file: tables/generative_tasks.tex
\begin{table}
    \centering
    \small
    \resizebox{0.7\textwidth}{!}{
    \begin{tabular}{lcccc}
        \toprule
        &  \multicolumn{3}{c}{Generative Tasks } \\
        \cmidrule{2-5}
        Model & \multicolumn{2}{c}{FLORES-200 (spBleu)} & XLSum (RougeL) \\
         & X$\rightarrow$ En & En $\rightarrow$ X & &\\
        \midrule
        \noalign{\smallskip} 
        Bactrian-X-7B & 25.9 & 16.6 & 7.7 \\
        Mistral-7B-Instruct-v0.2 & 31.1 & 21.0 & 6.3 \\
        Gemma-1.1-7B-it & 32.0 & 25.6 & 13.0 \\
        \aya-101-13B & 35.9 & 30.4 & \textbf{27.5} \\
        \includegraphics[scale=0.08]{./figures/logo2.png}\aya-23-8B & \textbf{39.5} & \textbf{34.8} & \textbf{27.5} \\
        \noalign{\smallskip} 
        \hdashline 
        \noalign{\smallskip}
        Mixtral-8x7B-Instruct-v0.1 & 36.3 & 28.9 & 7.1 \\
        \includegraphics[scale=0.08]{./figures/logo2.png}\aya-23-35B & \textbf{43.0} & \textbf{37.8} & \textbf{30.9}\\
        \bottomrule
    \end{tabular}
    }
    \caption{\textbf{Translation (FLORES)} and \textbf{multilingual summarization (XLSum)} results for baselines and \aya models. For XLSUM, we evaluate models on 15 languages that are included in \aya 23, and for FLORES we use all 22 languages paired with English. 
    }
    \label{tab:generative-results}
\end{table}

%% file: main.bbl
\begin{thebibliography}{63}
\providecommand{\natexlab}[1]{#1}
\providecommand{\url}[1]{\texttt{#1}}
\expandafter\ifx\csname urlstyle\endcsname\relax
  \providecommand{\doi}[1]{doi: #1}\else
  \providecommand{\doi}{doi: \begingroup \urlstyle{rm}\Url}\fi

\bibitem[eth(2023)]{ethnologue}
Ethnologue.
\newblock \url{https://www.ethnologue.com/insights/how-many-languages/}, 2023.
\newblock Accessed: 2023-06-17.

\bibitem[Adda et~al.(2016)Adda, Stüker, Adda-Decker, Ambouroue, Besacier, Blachon, Bonneau-Maynard, Godard, Hamlaoui, Idiatov, Kouarata, Lamel, Makasso, Rialland, {Van de Velde}, Yvon, and Zerbian]{ADDA20168}
Gilles Adda, Sebastian Stüker, Martine Adda-Decker, Odette Ambouroue, Laurent Besacier, David Blachon, Hélène Bonneau-Maynard, Pierre Godard, Fatima Hamlaoui, Dmitry Idiatov, Guy-Noël Kouarata, Lori Lamel, Emmanuel-Moselly Makasso, Annie Rialland, Mark {Van de Velde}, François Yvon, and Sabine Zerbian.
\newblock Breaking the unwritten language barrier: The bulb project.
\newblock \emph{Procedia Computer Science}, 81:\penalty0 8--14, 2016.
\newblock ISSN 1877-0509.
\newblock \doi{https://doi.org/10.1016/j.procs.2016.04.023}.
\newblock URL \url{https://www.sciencedirect.com/science/article/pii/S1877050916300370}.
\newblock SLTU-2016 5th Workshop on Spoken Language Technologies for Under-resourced languages 09-12 May 2016 Yogyakarta, Indonesia.

\bibitem[Ahia et~al.(2023)Ahia, Kumar, Gonen, Kasai, Mortensen, Smith, and Tsvetkov]{ahia2023languages}
Orevaoghene Ahia, Sachin Kumar, Hila Gonen, Jungo Kasai, David~R. Mortensen, Noah~A. Smith, and Yulia Tsvetkov.
\newblock Do all languages cost the same? tokenization in the era of commercial language models, 2023.

\bibitem[Ainslie et~al.(2023)Ainslie, Lee-Thorp, de~Jong, Zemlyanskiy, Lebrón, and Sanghai]{gqa}
Joshua Ainslie, James Lee-Thorp, Michiel de~Jong, Yury Zemlyanskiy, Federico Lebrón, and Sumit Sanghai.
\newblock Gqa: Training generalized multi-query transformer models from multi-head checkpoints, 2023.

\bibitem[Anil et~al.(2023)Anil, Dai, Firat, Johnson, Lepikhin, Passos, Shakeri, Taropa, Bailey, Chen, Chu, Clark, Shafey, Huang, Meier-Hellstern, Mishra, Moreira, Omernick, Robinson, Ruder, Tay, Xiao, Xu, Zhang, Abrego, Ahn, Austin, Barham, Botha, Bradbury, Brahma, Brooks, Catasta, Cheng, Cherry, Choquette-Choo, Chowdhery, Crepy, Dave, Dehghani, Dev, Devlin, Díaz, Du, Dyer, Feinberg, Feng, Fienber, Freitag, Garcia, Gehrmann, Gonzalez, Gur-Ari, Hand, Hashemi, Hou, Howland, Hu, Hui, Hurwitz, Isard, Ittycheriah, Jagielski, Jia, Kenealy, Krikun, Kudugunta, Lan, Lee, Lee, Li, Li, Li, Li, Li, Lim, Lin, Liu, Liu, Maggioni, Mahendru, Maynez, Misra, Moussalem, Nado, Nham, Ni, Nystrom, Parrish, Pellat, Polacek, Polozov, Pope, Qiao, Reif, Richter, Riley, Ros, Roy, Saeta, Samuel, Shelby, Slone, Smilkov, So, Sohn, Tokumine, Valter, Vasudevan, Vodrahalli, Wang, Wang, Wang, Wang, Wieting, Wu, Xu, Xu, Xue, Yin, Yu, Zhang, Zheng, Zheng, Zhou, Zhou, Petrov, and Wu]{anil2023palm}
Rohan Anil, Andrew~M. Dai, Orhan Firat, Melvin Johnson, Dmitry Lepikhin, Alexandre Passos, Siamak Shakeri, Emanuel Taropa, Paige Bailey, Zhifeng Chen, Eric Chu, Jonathan~H. Clark, Laurent~El Shafey, Yanping Huang, Kathy Meier-Hellstern, Gaurav Mishra, Erica Moreira, Mark Omernick, Kevin Robinson, Sebastian Ruder, Yi~Tay, Kefan Xiao, Yuanzhong Xu, Yujing Zhang, Gustavo~Hernandez Abrego, Junwhan Ahn, Jacob Austin, Paul Barham, Jan Botha, James Bradbury, Siddhartha Brahma, Kevin Brooks, Michele Catasta, Yong Cheng, Colin Cherry, Christopher~A. Choquette-Choo, Aakanksha Chowdhery, Clément Crepy, Shachi Dave, Mostafa Dehghani, Sunipa Dev, Jacob Devlin, Mark Díaz, Nan Du, Ethan Dyer, Vlad Feinberg, Fangxiaoyu Feng, Vlad Fienber, Markus Freitag, Xavier Garcia, Sebastian Gehrmann, Lucas Gonzalez, Guy Gur-Ari, Steven Hand, Hadi Hashemi, Le~Hou, Joshua Howland, Andrea Hu, Jeffrey Hui, Jeremy Hurwitz, Michael Isard, Abe Ittycheriah, Matthew Jagielski, Wenhao Jia, Kathleen Kenealy, Maxim Krikun, Sneha Kudugunta, Chang
  Lan, Katherine Lee, Benjamin Lee, Eric Li, Music Li, Wei Li, YaGuang Li, Jian Li, Hyeontaek Lim, Hanzhao Lin, Zhongtao Liu, Frederick Liu, Marcello Maggioni, Aroma Mahendru, Joshua Maynez, Vedant Misra, Maysam Moussalem, Zachary Nado, John Nham, Eric Ni, Andrew Nystrom, Alicia Parrish, Marie Pellat, Martin Polacek, Alex Polozov, Reiner Pope, Siyuan Qiao, Emily Reif, Bryan Richter, Parker Riley, Alex~Castro Ros, Aurko Roy, Brennan Saeta, Rajkumar Samuel, Renee Shelby, Ambrose Slone, Daniel Smilkov, David~R. So, Daniel Sohn, Simon Tokumine, Dasha Valter, Vijay Vasudevan, Kiran Vodrahalli, Xuezhi Wang, Pidong Wang, Zirui Wang, Tao Wang, John Wieting, Yuhuai Wu, Kelvin Xu, Yunhan Xu, Linting Xue, Pengcheng Yin, Jiahui Yu, Qiao Zhang, Steven Zheng, Ce~Zheng, Weikang Zhou, Denny Zhou, Slav Petrov, and Yonghui Wu.
\newblock Palm 2 technical report.
\newblock \emph{arXiv}, abs/2305.10403, 2023.

\bibitem[Arivazhagan et~al.(2019)Arivazhagan, Bapna, Firat, Lepikhin, Johnson, Krikun, Chen, Cao, Foster, Cherry, et~al.]{arivazhagan2019massively}
Naveen Arivazhagan, Ankur Bapna, Orhan Firat, Dmitry Lepikhin, Melvin Johnson, Maxim Krikun, Mia~Xu Chen, Yuan Cao, George Foster, Colin Cherry, et~al.
\newblock Massively multilingual neural machine translation in the wild: Findings and challenges.
\newblock \emph{arXiv preprint arXiv:1907.05019}, 2019.

\bibitem[Beeching et~al.(2023)Beeching, Fourrier, Habib, Han, Lambert, Rajani, Sanseviero, Tunstall, and Wolf]{open-llm-leaderboard}
Edward Beeching, Clémentine Fourrier, Nathan Habib, Sheon Han, Nathan Lambert, Nazneen Rajani, Omar Sanseviero, Lewis Tunstall, and Thomas Wolf.
\newblock Open llm leaderboard.
\newblock \url{https://huggingface.co/spaces/HuggingFaceH4/open_llm_leaderboard}, 2023.

\bibitem[Cobbe et~al.(2021)Cobbe, Kosaraju, Bavarian, Chen, Jun, Kaiser, Plappert, Tworek, Hilton, Nakano, et~al.]{cobbe2021training}
Karl Cobbe, Vineet Kosaraju, Mohammad Bavarian, Mark Chen, Heewoo Jun, Lukasz Kaiser, Matthias Plappert, Jerry Tworek, Jacob Hilton, Reiichiro Nakano, et~al.
\newblock Training verifiers to solve math word problems.
\newblock \emph{arXiv preprint arXiv:2110.14168}, 2021.

\bibitem[Conneau et~al.(2018)Conneau, Lample, Rinott, Williams, Bowman, Schwenk, and Stoyanov]{conneau2018xnli}
Alexis Conneau, Guillaume Lample, Ruty Rinott, Adina Williams, Samuel~R Bowman, Holger Schwenk, and Veselin Stoyanov.
\newblock Xnli: Evaluating cross-lingual sentence representations.
\newblock pp.\  2475--2485, October-November 2018.
\newblock \doi{10.18653/v1/D18-1269}.
\newblock URL \url{https://aclanthology.org/D18-1269}.

\bibitem[Conneau et~al.(2019)Conneau, Khandelwal, Goyal, Chaudhary, Wenzek, Guzm{\'a}n, Grave, Ott, Zettlemoyer, and Stoyanov]{conneau2019unsupervised}
Alexis Conneau, Kartikay Khandelwal, Naman Goyal, Vishrav Chaudhary, Guillaume Wenzek, Francisco Guzm{\'a}n, Edouard Grave, Myle Ott, Luke Zettlemoyer, and Veselin Stoyanov.
\newblock Unsupervised cross-lingual representation learning at scale.
\newblock pp.\  8440--8451, July 2019.
\newblock \doi{10.18653/v1/2020.acl-main.747}.
\newblock URL \url{https://aclanthology.org/2020.acl-main.747}.

\bibitem[Conover et~al.(2023{\natexlab{a}})Conover, Hayes, Mathur, Meng, Xie, Wan, Shah, Ghodsi, Wendell, Zaharia, et~al.]{conover2023free}
Mike Conover, Matt Hayes, Ankit Mathur, Xiangrui Meng, Jianwei Xie, Jun Wan, Sam Shah, Ali Ghodsi, Patrick Wendell, Matei Zaharia, et~al.
\newblock Free dolly: Introducing the world’s first truly open instruction-tuned llm.
\newblock \emph{Databricks}, 2023{\natexlab{a}}.

\bibitem[Conover et~al.(2023{\natexlab{b}})Conover, Hayes, Mathur, Xie, Wan, Shah, Ghodsi, Wendell, Zaharia, and Xin]{DatabricksBlog2023DollyV2}
Mike Conover, Matt Hayes, Ankit Mathur, Jianwei Xie, Jun Wan, Sam Shah, Ali Ghodsi, Patrick Wendell, Matei Zaharia, and Reynold Xin.
\newblock Free dolly: Introducing the world's first truly open instruction-tuned llm, 2023{\natexlab{b}}.
\newblock URL \url{https://www.databricks.com/blog/2023/04/12/dolly-first-open-commercially-viable-instruction-tuned-llm}.

\bibitem[Dac~Lai et~al.(2023)Dac~Lai, Van~Nguyen, Ngo, Nguyen, Dernoncourt, Rossi, and Nguyen]{dac2023okapi}
Viet Dac~Lai, Chien Van~Nguyen, Nghia~Trung Ngo, Thuat Nguyen, Franck Dernoncourt, Ryan~A Rossi, and Thien~Huu Nguyen.
\newblock Okapi: Instruction-tuned large language models in multiple languages with reinforcement learning from human feedback.
\newblock \emph{arXiv e-prints}, pp.\  arXiv--2307, 2023.

\bibitem[Deng et~al.(2023)Deng, Zhang, Pan, and Bing]{deng2023multilingual}
Yue Deng, Wenxuan Zhang, Sinno~Jialin Pan, and Lidong Bing.
\newblock Multilingual jailbreak challenges in large language models.
\newblock \emph{arXiv preprint arXiv:2310.06474}, 2023.

\bibitem[Dubois et~al.(2023)Dubois, Li, Taori, Zhang, Gulrajani, Ba, Guestrin, Liang, and Hashimoto]{dubois2023alpacafarm}
Yann Dubois, Xuechen Li, Rohan Taori, Tianyi Zhang, Ishaan Gulrajani, Jimmy Ba, Carlos Guestrin, Percy Liang, and Tatsunori~B Hashimoto.
\newblock Alpacafarm: A simulation framework for methods that learn from human feedback.
\newblock \emph{arXiv preprint arXiv:2305.14387}, 2023.

\bibitem[Durmus et~al.(2023)Durmus, Nyugen, Liao, Schiefer, Askell, Bakhtin, Chen, Hatfield-Dodds, Hernandez, Joseph, Lovitt, McCandlish, Sikder, Tamkin, Thamkul, Kaplan, Clark, and Ganguli]{durmus2023measuring}
Esin Durmus, Karina Nyugen, Thomas~I. Liao, Nicholas Schiefer, Amanda Askell, Anton Bakhtin, Carol Chen, Zac Hatfield-Dodds, Danny Hernandez, Nicholas Joseph, Liane Lovitt, Sam McCandlish, Orowa Sikder, Alex Tamkin, Janel Thamkul, Jared Kaplan, Jack Clark, and Deep Ganguli.
\newblock Towards measuring the representation of subjective global opinions in language models.
\newblock \emph{arXiv}, abs/2306.16388, 2023.

\bibitem[Gao et~al.(2023)Gao, Tow, Abbasi, Biderman, Black, DiPofi, Foster, Golding, Hsu, Le~Noac'h, Li, McDonell, Muennighoff, Ociepa, Phang, Reynolds, Schoelkopf, Skowron, Sutawika, Tang, Thite, Wang, Wang, and Zou]{eval-harness}
Leo Gao, Jonathan Tow, Baber Abbasi, Stella Biderman, Sid Black, Anthony DiPofi, Charles Foster, Laurence Golding, Jeffrey Hsu, Alain Le~Noac'h, Haonan Li, Kyle McDonell, Niklas Muennighoff, Chris Ociepa, Jason Phang, Laria Reynolds, Hailey Schoelkopf, Aviya Skowron, Lintang Sutawika, Eric Tang, Anish Thite, Ben Wang, Kevin Wang, and Andy Zou.
\newblock A framework for few-shot language model evaluation.
\newblock 12 2023.
\newblock \doi{10.5281/zenodo.10256836}.
\newblock URL \url{https://zenodo.org/records/10256836}.

\bibitem[Gemini-Team et~al.(2024)Gemini-Team, Anil, Borgeaud, Alayrac, Yu, Soricut, Schalkwyk, Dai, Hauth, Millican, Silver, Johnson, Antonoglou, Schrittwieser, Glaese, Chen, Pitler, Lillicrap, Lazaridou, Firat, Molloy, Isard, Barham, Hennigan, Lee, Viola, Reynolds, Xu, Doherty, Collins, Meyer, Rutherford, Moreira, Ayoub, Goel, Krawczyk, Du, Chi, Cheng, Ni, Shah, Kane, Chan, Faruqui, Severyn, Lin, Li, Cheng, Ittycheriah, Mahdieh, Chen, Sun, Tran, Bagri, Lakshminarayanan, Liu, Orban, Güra, Zhou, Song, Boffy, Ganapathy, Zheng, Choe, Ágoston Weisz, Zhu, Lu, Gopal, Kahn, Kula, Pitman, Shah, Taropa, Merey, Baeuml, Chen, Shafey, Zhang, Sercinoglu, Tucker, Piqueras, Krikun, Barr, Savinov, Danihelka, Roelofs, White, Andreassen, von Glehn, Yagati, Kazemi, Gonzalez, Khalman, Sygnowski, Frechette, Smith, Culp, Proleev, Luan, Chen, Lottes, Schucher, Lebron, Rrustemi, Clay, Crone, Kocisky, Zhao, Perz, Yu, Howard, Bloniarz, Rae, Lu, Sifre, Maggioni, Alcober, Garrette, Barnes, Thakoor, Austin, Barth-Maron, Wong, Joshi,
  Chaabouni, Fatiha, Ahuja, Tomar, Senter, Chadwick, Kornakov, Attaluri, Iturrate, Liu, Li, Cogan, Chen, Jia, Gu, Zhang, Grimstad, Hartman, Garcia, Pillai, Devlin, Laskin, de~Las~Casas, Valter, Tao, Blanco, Badia, Reitter, Chen, Brennan, Rivera, Brin, Iqbal, Surita, Labanowski, Rao, Winkler, Parisotto, Gu, Olszewska, Addanki, Miech, Louis, Teplyashin, Brown, Catt, Balaguer, Xiang, Wang, Ashwood, Briukhov, Webson, Ganapathy, Sanghavi, Kannan, Chang, Stjerngren, Djolonga, Sun, Bapna, Aitchison, Pejman, Michalewski, Yu, Wang, Love, Ahn, Bloxwich, Han, Humphreys, Sellam, Bradbury, Godbole, Samangooei, Damoc, Kaskasoli, Arnold, Vasudevan, Agrawal, Riesa, Lepikhin, Tanburn, Srinivasan, Lim, Hodkinson, Shyam, Ferret, Hand, Garg, Paine, Li, Li, Giang, Neitz, Abbas, York, Reid, Cole, Chowdhery, Das, Rogozińska, Nikolaev, Sprechmann, Nado, Zilka, Prost, He, Monteiro, Mishra, Welty, Newlan, Jia, Allamanis, Hu, de~Liedekerke, Gilmer, Saroufim, Rijhwani, Hou, Shrivastava, Baddepudi, Goldin, Ozturel, Cassirer, Xu, Sohn,
  Sachan, Amplayo, Swanson, Petrova, Narayan, Guez, Brahma, Landon, Patel, Zhao, Villela, Wang, Jia, Rahtz, Giménez, Yeung, Keeling, Georgiev, Mincu, Wu, Haykal, Saputro, Vodrahalli, Qin, Cankara, Sharma, Fernando, Hawkins, Neyshabur, Kim, Hutter, Agrawal, Castro-Ros, van~den Driessche, Wang, Yang, yiin Chang, Komarek, McIlroy, Lučić, Zhang, Farhan, Sharman, Natsev, Michel, Bansal, Qiao, Cao, Shakeri, Butterfield, Chung, Rubenstein, Agrawal, Mensch, Soparkar, Lenc, Chung, Pope, Maggiore, Kay, Jhakra, Wang, Maynez, Phuong, Tobin, Tacchetti, Trebacz, Robinson, Katariya, Riedel, Bailey, Xiao, Ghelani, Aroyo, Slone, Houlsby, Xiong, Yang, Gribovskaya, Adler, Wirth, Lee, Li, Kagohara, Pavagadhi, Bridgers, Bortsova, Ghemawat, Ahmed, Liu, Powell, Bolina, Iinuma, Zablotskaia, Besley, Chung, Dozat, Comanescu, Si, Greer, Su, Polacek, Kaufman, Tokumine, Hu, Buchatskaya, Miao, Elhawaty, Siddhant, Tomasev, Xing, Greer, Miller, Ashraf, Roy, Zhang, Ma, Filos, Besta, Blevins, Klimenko, Yeh, Changpinyo, Mu, Chang,
  Pajarskas, Muir, Cohen, Lan, Haridasan, Marathe, Hansen, Douglas, Samuel, Wang, Austin, Lan, Jiang, Chiu, Lorenzo, Sjösund, Cevey, Gleicher, Avrahami, Boral, Srinivasan, Selo, May, Aisopos, Hussenot, Soares, Baumli, Chang, Recasens, Caine, Pritzel, Pavetic, Pardo, Gergely, Frye, Ramasesh, Horgan, Badola, Kassner, Roy, Dyer, Campos, Tomala, Tang, Badawy, White, Mustafa, Lang, Jindal, Vikram, Gong, Caelles, Hemsley, Thornton, Feng, Stokowiec, Zheng, Thacker, Çağlar Ünlü, Zhang, Saleh, Svensson, Bileschi, Patil, Anand, Ring, Tsihlas, Vezer, Selvi, Shevlane, Rodriguez, Kwiatkowski, Daruki, Rong, Dafoe, FitzGerald, Gu-Lemberg, Khan, Hendricks, Pellat, Feinberg, Cobon-Kerr, Sainath, Rauh, Hashemi, Ives, Hasson, Noland, Cao, Byrd, Hou, Wang, Sottiaux, Paganini, Lespiau, Moufarek, Hassan, Shivakumar, van Amersfoort, Mandhane, Joshi, Goyal, Tung, Brock, Sheahan, Misra, Li, Rakićević, Dehghani, Liu, Mittal, Oh, Noury, Sezener, Huot, Lamm, Cao, Chen, Mudgal, Stella, Brooks, Vasudevan, Liu, Chain, Melinkeri,
  Cohen, Wang, Seymore, Zubkov, Goel, Yue, Krishnakumaran, Albert, Hurley, Sano, Mohananey, Joughin, Filonov, Kępa, Eldawy, Lim, Rishi, Badiezadegan, Bos, Chang, Jain, Padmanabhan, Puttagunta, Krishna, Baker, Kalb, Bedapudi, Kurzrok, Lei, Yu, Litvin, Zhou, Wu, Sobell, Siciliano, Papir, Neale, Bragagnolo, Toor, Chen, Anklin, Wang, Feng, Gholami, Ling, Liu, Walter, Moghaddam, Kishore, Adamek, Mercado, Mallinson, Wandekar, Cagle, Ofek, Garrido, Lombriser, Mukha, Sun, Mohammad, Matak, Qian, Peswani, Janus, Yuan, Schelin, David, Garg, He, Duzhyi, Älgmyr, Lottaz, Li, Yadav, Xu, Chinien, Shivanna, Chuklin, Li, Spadine, Wolfe, Mohamed, Das, Dai, He, von Dincklage, Upadhyay, Maurya, Chi, Krause, Salama, Rabinovitch, M, Selvan, Dektiarev, Ghiasi, Guven, Gupta, Liu, Sharma, Shtacher, Paul, Akerlund, Aubet, Huang, Zhu, Zhu, Teixeira, Fritze, Bertolini, Marinescu, Bölle, Paulus, Gupta, Latkar, Chang, Sanders, Wilson, Wu, Tan, Thiet, Doshi, Lall, Mishra, Chen, Luong, Benjamin, Lee, Andrejczuk, Rabiej, Ranjan, Styrc,
  Yin, Simon, Harriott, Bansal, Robsky, Bacon, Greene, Mirylenka, Zhou, Sarvana, Goyal, Andermatt, Siegler, Horn, Israel, Pongetti, Chen, Selvatici, Silva, Wang, Tolins, Guu, Yogev, Cai, Agostini, Shah, Nguyen, Donnaile, Pereira, Friso, Stambler, Kurzrok, Kuang, Romanikhin, Geller, Yan, Jang, Lee, Fica, Malmi, Tan, Banica, Balle, Pham, Huang, Avram, Shi, Singh, Hidey, Ahuja, Saxena, Dooley, Potharaju, O'Neill, Gokulchandran, Foley, Zhao, Dusenberry, Liu, Mehta, Kotikalapudi, Safranek-Shrader, Goodman, Kessinger, Globen, Kolhar, Gorgolewski, Ibrahim, Song, Eichenbaum, Brovelli, Potluri, Lahoti, Baetu, Ghorbani, Chen, Crawford, Pal, Sridhar, Gurita, Mujika, Petrovski, Cedoz, Li, Chen, Santo, Goyal, Punjabi, Kappaganthu, Kwak, LV, Velury, Choudhury, Hall, Shah, Figueira, Thomas, Lu, Zhou, Kumar, Jurdi, Chikkerur, Ma, Yu, Kwak, Ähdel, Rajayogam, Choma, Liu, Barua, Ji, Park, Hellendoorn, Bailey, Bilal, Zhou, Khatir, Sutton, Rzadkowski, Macintosh, Shagin, Medina, Liang, Zhou, Shah, Bi, Dankovics, Banga, Lehmann,
  Bredesen, Lin, Hoffmann, Lai, Chung, Yang, Balani, Bražinskas, Sozanschi, Hayes, Alcalde, Makarov, Chen, Stella, Snijders, Mandl, Kärrman, Nowak, Wu, Dyck, Vaidyanathan, R, Mallet, Rudominer, Johnston, Mittal, Udathu, Christensen, Verma, Irving, Santucci, Elsayed, Davoodi, Georgiev, Tenney, Hua, Cideron, Leurent, Alnahlawi, Georgescu, Wei, Zheng, Scandinaro, Jiang, Snoek, Sundararajan, Wang, Ontiveros, Karo, Cole, Rajashekhar, Tumeh, Ben-David, Jain, Uesato, Datta, Bunyan, Wu, Zhang, Stanczyk, Zhang, Steiner, Naskar, Azzam, Johnson, Paszke, Chiu, Elias, Mohiuddin, Muhammad, Miao, Lee, Vieillard, Park, Zhang, Stanway, Garmon, Karmarkar, Dong, Lee, Kumar, Zhou, Evens, Isaac, Irving, Loper, Fink, Arkatkar, Chen, Shafran, Petrychenko, Chen, Jia, Levskaya, Zhu, Grabowski, Mao, Magni, Yao, Snaider, Casagrande, Palmer, Suganthan, Castaño, Giannoumis, Kim, Rybiński, Sreevatsa, Prendki, Soergel, Goedeckemeyer, Gierke, Jafari, Gaba, Wiesner, Wright, Wei, Vashisht, Kulizhskaya, Hoover, Le, Li, Iwuanyanwu, Liu,
  Ramirez, Khorlin, Cui, LIN, Wu, Aguilar, Pallo, Chakladar, Perng, Abellan, Zhang, Dasgupta, Kushman, Penchev, Repina, Wu, van~der Weide, Ponnapalli, Kaplan, Simsa, Li, Dousse, Yang, Piper, Ie, Pasumarthi, Lintz, Vijayakumar, Andor, Valenzuela, Lui, Paduraru, Peng, Lee, Zhang, Greene, Nguyen, Kurylowicz, Hardin, Dixon, Janzer, Choo, Feng, Zhang, Singhal, Du, McKinnon, Antropova, Bolukbasi, Keller, Reid, Finchelstein, Raad, Crocker, Hawkins, Dadashi, Gaffney, Franko, Bulanova, Leblond, Chung, Askham, Cobo, Xu, Fischer, Xu, Sorokin, Alberti, Lin, Evans, Dimitriev, Forbes, Banarse, Tung, Omernick, Bishop, Sterneck, Jain, Xia, Amid, Piccinno, Wang, Banzal, Mankowitz, Polozov, Krakovna, Brown, Bateni, Duan, Firoiu, Thotakuri, Natan, Geist, tan Girgin, Li, Ye, Roval, Tojo, Kwong, Lee-Thorp, Yew, Sinopalnikov, Ramos, Mellor, Sharma, Wu, Miller, Sonnerat, Vnukov, Greig, Beattie, Caveness, Bai, Eisenschlos, Korchemniy, Tsai, Jasarevic, Kong, Dao, Zheng, Liu, Yang, Zhu, Teh, Sanmiya, Gladchenko, Trdin, Toyama, Rosen,
  Tavakkol, Xue, Elkind, Woodman, Carpenter, Papamakarios, Kemp, Kafle, Grunina, Sinha, Talbert, Wu, Owusu-Afriyie, Du, Thornton, Pont-Tuset, Narayana, Li, Fatehi, Wieting, Ajmeri, Uria, Ko, Knight, Héliou, Niu, Gu, Pang, Li, Levine, Stolovich, Santamaria-Fernandez, Goenka, Yustalim, Strudel, Elqursh, Deck, Lee, Li, Levin, Hoffmann, Holtmann-Rice, Bachem, Arora, Koh, Yeganeh, Põder, Tariq, Sun, Ionita, Seyedhosseini, Tafti, Liu, Gulati, Liu, Ye, Chrzaszcz, Wang, Sethi, Li, Brown, Singh, Fan, Parisi, Stanton, Koverkathu, Choquette-Choo, Li, Lu, Ittycheriah, Shroff, Varadarajan, Bahargam, Willoughby, Gaddy, Desjardins, Cornero, Robenek, Mittal, Albrecht, Shenoy, Moiseev, Jacobsson, Ghaffarkhah, Rivière, Walton, Crepy, Parrish, Zhou, Farabet, Radebaugh, Srinivasan, van~der Salm, Fidjeland, Scellato, Latorre-Chimoto, Klimczak-Plucińska, Bridson, de~Cesare, Hudson, Mendolicchio, Walker, Morris, Mauger, Guseynov, Reid, Odoom, Loher, Cotruta, Yenugula, Grewe, Petrushkina, Duerig, Sanchez, Yadlowsky, Shen,
  Globerson, Webb, Dua, Li, Bhupatiraju, Hurt, Qureshi, Agarwal, Shani, Eyal, Khare, Belle, Wang, Tekur, Kale, Wei, Sang, Saeta, Liechty, Sun, Zhao, Lee, Nayak, Fritz, Vuyyuru, Aslanides, Vyas, Wicke, Ma, Eltyshev, Martin, Cate, Manyika, Amiri, Kim, Xiong, Kang, Luisier, Tripuraneni, Madras, Guo, Waters, Wang, Ainslie, Baldridge, Zhang, Pruthi, Bauer, Yang, Mansour, Gelman, Xu, Polovets, Liu, Cai, Chen, Sheng, Xue, Ozair, Angermueller, Li, Sinha, Wang, Wiesinger, Koukoumidis, Tian, Iyer, Gurumurthy, Goldenson, Shah, Blake, Yu, Urbanowicz, Palomaki, Fernando, Durden, Mehta, Momchev, Rahimtoroghi, Georgaki, Raul, Ruder, Redshaw, Lee, Zhou, Jalan, Li, Hechtman, Schuh, Nasr, Milan, Mikulik, Franco, Green, Nguyen, Kelley, Mahendru, Hu, Howland, Vargas, Hui, Bansal, Rao, Ghiya, Wang, Ye, Sarr, Preston, Elish, Li, Kaku, Gupta, Pasupat, Juan, Someswar, M., Chen, Amini, Fabrikant, Chu, Dong, Muthal, Buthpitiya, Jauhari, Hua, Khandelwal, Hitron, Ren, Rinaldi, Drath, Dabush, Jiang, Godhia, Sachs, Chen, Fan, Taitelbaum,
  Noga, Dai, Wang, Liang, Hamer, Ferng, Elkind, Atias, Lee, Listík, Carlen, van~de Kerkhof, Pikus, Zaher, Müller, Zykova, Stefanec, Gatsko, Hirnschall, Sethi, Xu, Ahuja, Tsai, Stefanoiu, Feng, Dhandhania, Katyal, Gupta, Parulekar, Pitta, Zhao, Bhatia, Bhavnani, Alhadlaq, Li, Danenberg, Tu, Pine, Filippova, Ghosh, Limonchik, Urala, Lanka, Clive, Sun, Li, Wu, Hongtongsak, Li, Thakkar, Omarov, Majmundar, Alverson, Kucharski, Patel, Jain, Zabelin, Pelagatti, Kohli, Kumar, Kim, Sankar, Shah, Ramachandruni, Zeng, Bariach, Weidinger, Subramanya, Hsiao, Hassabis, Kavukcuoglu, Sadovsky, Le, Strohman, Wu, Petrov, Dean, and Vinyals]{geminiteam2024gemini}
Gemini-Team, Rohan Anil, Sebastian Borgeaud, Jean-Baptiste Alayrac, Jiahui Yu, Radu Soricut, Johan Schalkwyk, Andrew~M. Dai, Anja Hauth, Katie Millican, David Silver, Melvin Johnson, Ioannis Antonoglou, Julian Schrittwieser, Amelia Glaese, Jilin Chen, Emily Pitler, Timothy Lillicrap, Angeliki Lazaridou, Orhan Firat, James Molloy, Michael Isard, Paul~R. Barham, Tom Hennigan, Benjamin Lee, Fabio Viola, Malcolm Reynolds, Yuanzhong Xu, Ryan Doherty, Eli Collins, Clemens Meyer, Eliza Rutherford, Erica Moreira, Kareem Ayoub, Megha Goel, Jack Krawczyk, Cosmo Du, Ed~Chi, Heng-Tze Cheng, Eric Ni, Purvi Shah, Patrick Kane, Betty Chan, Manaal Faruqui, Aliaksei Severyn, Hanzhao Lin, YaGuang Li, Yong Cheng, Abe Ittycheriah, Mahdis Mahdieh, Mia Chen, Pei Sun, Dustin Tran, Sumit Bagri, Balaji Lakshminarayanan, Jeremiah Liu, Andras Orban, Fabian Güra, Hao Zhou, Xinying Song, Aurelien Boffy, Harish Ganapathy, Steven Zheng, HyunJeong Choe, Ágoston Weisz, Tao Zhu, Yifeng Lu, Siddharth Gopal, Jarrod Kahn, Maciej Kula, Jeff
  Pitman, Rushin Shah, Emanuel Taropa, Majd~Al Merey, Martin Baeuml, Zhifeng Chen, Laurent~El Shafey, Yujing Zhang, Olcan Sercinoglu, George Tucker, Enrique Piqueras, Maxim Krikun, Iain Barr, Nikolay Savinov, Ivo Danihelka, Becca Roelofs, Anaïs White, Anders Andreassen, Tamara von Glehn, Lakshman Yagati, Mehran Kazemi, Lucas Gonzalez, Misha Khalman, Jakub Sygnowski, Alexandre Frechette, Charlotte Smith, Laura Culp, Lev Proleev, Yi~Luan, Xi~Chen, James Lottes, Nathan Schucher, Federico Lebron, Alban Rrustemi, Natalie Clay, Phil Crone, Tomas Kocisky, Jeffrey Zhao, Bartek Perz, Dian Yu, Heidi Howard, Adam Bloniarz, Jack~W. Rae, Han Lu, Laurent Sifre, Marcello Maggioni, Fred Alcober, Dan Garrette, Megan Barnes, Shantanu Thakoor, Jacob Austin, Gabriel Barth-Maron, William Wong, Rishabh Joshi, Rahma Chaabouni, Deeni Fatiha, Arun Ahuja, Gaurav~Singh Tomar, Evan Senter, Martin Chadwick, Ilya Kornakov, Nithya Attaluri, Iñaki Iturrate, Ruibo Liu, Yunxuan Li, Sarah Cogan, Jeremy Chen, Chao Jia, Chenjie Gu, Qiao Zhang,
  Jordan Grimstad, Ale~Jakse Hartman, Xavier Garcia, Thanumalayan~Sankaranarayana Pillai, Jacob Devlin, Michael Laskin, Diego de~Las~Casas, Dasha Valter, Connie Tao, Lorenzo Blanco, Adrià~Puigdomènech Badia, David Reitter, Mianna Chen, Jenny Brennan, Clara Rivera, Sergey Brin, Shariq Iqbal, Gabriela Surita, Jane Labanowski, Abhi Rao, Stephanie Winkler, Emilio Parisotto, Yiming Gu, Kate Olszewska, Ravi Addanki, Antoine Miech, Annie Louis, Denis Teplyashin, Geoff Brown, Elliot Catt, Jan Balaguer, Jackie Xiang, Pidong Wang, Zoe Ashwood, Anton Briukhov, Albert Webson, Sanjay Ganapathy, Smit Sanghavi, Ajay Kannan, Ming-Wei Chang, Axel Stjerngren, Josip Djolonga, Yuting Sun, Ankur Bapna, Matthew Aitchison, Pedram Pejman, Henryk Michalewski, Tianhe Yu, Cindy Wang, Juliette Love, Junwhan Ahn, Dawn Bloxwich, Kehang Han, Peter Humphreys, Thibault Sellam, James Bradbury, Varun Godbole, Sina Samangooei, Bogdan Damoc, Alex Kaskasoli, Sébastien M.~R. Arnold, Vijay Vasudevan, Shubham Agrawal, Jason Riesa, Dmitry
  Lepikhin, Richard Tanburn, Srivatsan Srinivasan, Hyeontaek Lim, Sarah Hodkinson, Pranav Shyam, Johan Ferret, Steven Hand, Ankush Garg, Tom~Le Paine, Jian Li, Yujia Li, Minh Giang, Alexander Neitz, Zaheer Abbas, Sarah York, Machel Reid, Elizabeth Cole, Aakanksha Chowdhery, Dipanjan Das, Dominika Rogozińska, Vitaliy Nikolaev, Pablo Sprechmann, Zachary Nado, Lukas Zilka, Flavien Prost, Luheng He, Marianne Monteiro, Gaurav Mishra, Chris Welty, Josh Newlan, Dawei Jia, Miltiadis Allamanis, Clara~Huiyi Hu, Raoul de~Liedekerke, Justin Gilmer, Carl Saroufim, Shruti Rijhwani, Shaobo Hou, Disha Shrivastava, Anirudh Baddepudi, Alex Goldin, Adnan Ozturel, Albin Cassirer, Yunhan Xu, Daniel Sohn, Devendra Sachan, Reinald~Kim Amplayo, Craig Swanson, Dessie Petrova, Shashi Narayan, Arthur Guez, Siddhartha Brahma, Jessica Landon, Miteyan Patel, Ruizhe Zhao, Kevin Villela, Luyu Wang, Wenhao Jia, Matthew Rahtz, Mai Giménez, Legg Yeung, James Keeling, Petko Georgiev, Diana Mincu, Boxi Wu, Salem Haykal, Rachel Saputro, Kiran
  Vodrahalli, James Qin, Zeynep Cankara, Abhanshu Sharma, Nick Fernando, Will Hawkins, Behnam Neyshabur, Solomon Kim, Adrian Hutter, Priyanka Agrawal, Alex Castro-Ros, George van~den Driessche, Tao Wang, Fan Yang, Shuo yiin Chang, Paul Komarek, Ross McIlroy, Mario Lučić, Guodong Zhang, Wael Farhan, Michael Sharman, Paul Natsev, Paul Michel, Yamini Bansal, Siyuan Qiao, Kris Cao, Siamak Shakeri, Christina Butterfield, Justin Chung, Paul~Kishan Rubenstein, Shivani Agrawal, Arthur Mensch, Kedar Soparkar, Karel Lenc, Timothy Chung, Aedan Pope, Loren Maggiore, Jackie Kay, Priya Jhakra, Shibo Wang, Joshua Maynez, Mary Phuong, Taylor Tobin, Andrea Tacchetti, Maja Trebacz, Kevin Robinson, Yash Katariya, Sebastian Riedel, Paige Bailey, Kefan Xiao, Nimesh Ghelani, Lora Aroyo, Ambrose Slone, Neil Houlsby, Xuehan Xiong, Zhen Yang, Elena Gribovskaya, Jonas Adler, Mateo Wirth, Lisa Lee, Music Li, Thais Kagohara, Jay Pavagadhi, Sophie Bridgers, Anna Bortsova, Sanjay Ghemawat, Zafarali Ahmed, Tianqi Liu, Richard Powell,
  Vijay Bolina, Mariko Iinuma, Polina Zablotskaia, James Besley, Da-Woon Chung, Timothy Dozat, Ramona Comanescu, Xiance Si, Jeremy Greer, Guolong Su, Martin Polacek, Raphaël~Lopez Kaufman, Simon Tokumine, Hexiang Hu, Elena Buchatskaya, Yingjie Miao, Mohamed Elhawaty, Aditya Siddhant, Nenad Tomasev, Jinwei Xing, Christina Greer, Helen Miller, Shereen Ashraf, Aurko Roy, Zizhao Zhang, Ada Ma, Angelos Filos, Milos Besta, Rory Blevins, Ted Klimenko, Chih-Kuan Yeh, Soravit Changpinyo, Jiaqi Mu, Oscar Chang, Mantas Pajarskas, Carrie Muir, Vered Cohen, Charline~Le Lan, Krishna Haridasan, Amit Marathe, Steven Hansen, Sholto Douglas, Rajkumar Samuel, Mingqiu Wang, Sophia Austin, Chang Lan, Jiepu Jiang, Justin Chiu, Jaime~Alonso Lorenzo, Lars~Lowe Sjösund, Sébastien Cevey, Zach Gleicher, Thi Avrahami, Anudhyan Boral, Hansa Srinivasan, Vittorio Selo, Rhys May, Konstantinos Aisopos, Léonard Hussenot, Livio~Baldini Soares, Kate Baumli, Michael~B. Chang, Adrià Recasens, Ben Caine, Alexander Pritzel, Filip Pavetic,
  Fabio Pardo, Anita Gergely, Justin Frye, Vinay Ramasesh, Dan Horgan, Kartikeya Badola, Nora Kassner, Subhrajit Roy, Ethan Dyer, Víctor~Campos Campos, Alex Tomala, Yunhao Tang, Dalia~El Badawy, Elspeth White, Basil Mustafa, Oran Lang, Abhishek Jindal, Sharad Vikram, Zhitao Gong, Sergi Caelles, Ross Hemsley, Gregory Thornton, Fangxiaoyu Feng, Wojciech Stokowiec, Ce~Zheng, Phoebe Thacker, Çağlar Ünlü, Zhishuai Zhang, Mohammad Saleh, James Svensson, Max Bileschi, Piyush Patil, Ankesh Anand, Roman Ring, Katerina Tsihlas, Arpi Vezer, Marco Selvi, Toby Shevlane, Mikel Rodriguez, Tom Kwiatkowski, Samira Daruki, Keran Rong, Allan Dafoe, Nicholas FitzGerald, Keren Gu-Lemberg, Mina Khan, Lisa~Anne Hendricks, Marie Pellat, Vladimir Feinberg, James Cobon-Kerr, Tara Sainath, Maribeth Rauh, Sayed~Hadi Hashemi, Richard Ives, Yana Hasson, Eric Noland, Yuan Cao, Nathan Byrd, Le~Hou, Qingze Wang, Thibault Sottiaux, Michela Paganini, Jean-Baptiste Lespiau, Alexandre Moufarek, Samer Hassan, Kaushik Shivakumar, Joost van
  Amersfoort, Amol Mandhane, Pratik Joshi, Anirudh Goyal, Matthew Tung, Andrew Brock, Hannah Sheahan, Vedant Misra, Cheng Li, Nemanja Rakićević, Mostafa Dehghani, Fangyu Liu, Sid Mittal, Junhyuk Oh, Seb Noury, Eren Sezener, Fantine Huot, Matthew Lamm, Nicola~De Cao, Charlie Chen, Sidharth Mudgal, Romina Stella, Kevin Brooks, Gautam Vasudevan, Chenxi Liu, Mainak Chain, Nivedita Melinkeri, Aaron Cohen, Venus Wang, Kristie Seymore, Sergey Zubkov, Rahul Goel, Summer Yue, Sai Krishnakumaran, Brian Albert, Nate Hurley, Motoki Sano, Anhad Mohananey, Jonah Joughin, Egor Filonov, Tomasz Kępa, Yomna Eldawy, Jiawern Lim, Rahul Rishi, Shirin Badiezadegan, Taylor Bos, Jerry Chang, Sanil Jain, Sri Gayatri~Sundara Padmanabhan, Subha Puttagunta, Kalpesh Krishna, Leslie Baker, Norbert Kalb, Vamsi Bedapudi, Adam Kurzrok, Shuntong Lei, Anthony Yu, Oren Litvin, Xiang Zhou, Zhichun Wu, Sam Sobell, Andrea Siciliano, Alan Papir, Robby Neale, Jonas Bragagnolo, Tej Toor, Tina Chen, Valentin Anklin, Feiran Wang, Richie Feng, Milad
  Gholami, Kevin Ling, Lijuan Liu, Jules Walter, Hamid Moghaddam, Arun Kishore, Jakub Adamek, Tyler Mercado, Jonathan Mallinson, Siddhinita Wandekar, Stephen Cagle, Eran Ofek, Guillermo Garrido, Clemens Lombriser, Maksim Mukha, Botu Sun, Hafeezul~Rahman Mohammad, Josip Matak, Yadi Qian, Vikas Peswani, Pawel Janus, Quan Yuan, Leif Schelin, Oana David, Ankur Garg, Yifan He, Oleksii Duzhyi, Anton Älgmyr, Timothée Lottaz, Qi~Li, Vikas Yadav, Luyao Xu, Alex Chinien, Rakesh Shivanna, Aleksandr Chuklin, Josie Li, Carrie Spadine, Travis Wolfe, Kareem Mohamed, Subhabrata Das, Zihang Dai, Kyle He, Daniel von Dincklage, Shyam Upadhyay, Akanksha Maurya, Luyan Chi, Sebastian Krause, Khalid Salama, Pam~G Rabinovitch, Pavan Kumar~Reddy M, Aarush Selvan, Mikhail Dektiarev, Golnaz Ghiasi, Erdem Guven, Himanshu Gupta, Boyi Liu, Deepak Sharma, Idan~Heimlich Shtacher, Shachi Paul, Oscar Akerlund, François-Xavier Aubet, Terry Huang, Chen Zhu, Eric Zhu, Elico Teixeira, Matthew Fritze, Francesco Bertolini, Liana-Eleonora
  Marinescu, Martin Bölle, Dominik Paulus, Khyatti Gupta, Tejasi Latkar, Max Chang, Jason Sanders, Roopa Wilson, Xuewei Wu, Yi-Xuan Tan, Lam~Nguyen Thiet, Tulsee Doshi, Sid Lall, Swaroop Mishra, Wanming Chen, Thang Luong, Seth Benjamin, Jasmine Lee, Ewa Andrejczuk, Dominik Rabiej, Vipul Ranjan, Krzysztof Styrc, Pengcheng Yin, Jon Simon, Malcolm~Rose Harriott, Mudit Bansal, Alexei Robsky, Geoff Bacon, David Greene, Daniil Mirylenka, Chen Zhou, Obaid Sarvana, Abhimanyu Goyal, Samuel Andermatt, Patrick Siegler, Ben Horn, Assaf Israel, Francesco Pongetti, Chih-Wei~"Louis" Chen, Marco Selvatici, Pedro Silva, Kathie Wang, Jackson Tolins, Kelvin Guu, Roey Yogev, Xiaochen Cai, Alessandro Agostini, Maulik Shah, Hung Nguyen, Noah~Ó Donnaile, Sébastien Pereira, Linda Friso, Adam Stambler, Adam Kurzrok, Chenkai Kuang, Yan Romanikhin, Mark Geller, ZJ~Yan, Kane Jang, Cheng-Chun Lee, Wojciech Fica, Eric Malmi, Qijun Tan, Dan Banica, Daniel Balle, Ryan Pham, Yanping Huang, Diana Avram, Hongzhi Shi, Jasjot Singh, Chris
  Hidey, Niharika Ahuja, Pranab Saxena, Dan Dooley, Srividya~Pranavi Potharaju, Eileen O'Neill, Anand Gokulchandran, Ryan Foley, Kai Zhao, Mike Dusenberry, Yuan Liu, Pulkit Mehta, Ragha Kotikalapudi, Chalence Safranek-Shrader, Andrew Goodman, Joshua Kessinger, Eran Globen, Prateek Kolhar, Chris Gorgolewski, Ali Ibrahim, Yang Song, Ali Eichenbaum, Thomas Brovelli, Sahitya Potluri, Preethi Lahoti, Cip Baetu, Ali Ghorbani, Charles Chen, Andy Crawford, Shalini Pal, Mukund Sridhar, Petru Gurita, Asier Mujika, Igor Petrovski, Pierre-Louis Cedoz, Chenmei Li, Shiyuan Chen, Niccolò~Dal Santo, Siddharth Goyal, Jitesh Punjabi, Karthik Kappaganthu, Chester Kwak, Pallavi LV, Sarmishta Velury, Himadri Choudhury, Jamie Hall, Premal Shah, Ricardo Figueira, Matt Thomas, Minjie Lu, Ting Zhou, Chintu Kumar, Thomas Jurdi, Sharat Chikkerur, Yenai Ma, Adams Yu, Soo Kwak, Victor Ähdel, Sujeevan Rajayogam, Travis Choma, Fei Liu, Aditya Barua, Colin Ji, Ji~Ho Park, Vincent Hellendoorn, Alex Bailey, Taylan Bilal, Huanjie Zhou,
  Mehrdad Khatir, Charles Sutton, Wojciech Rzadkowski, Fiona Macintosh, Konstantin Shagin, Paul Medina, Chen Liang, Jinjing Zhou, Pararth Shah, Yingying Bi, Attila Dankovics, Shipra Banga, Sabine Lehmann, Marissa Bredesen, Zifan Lin, John~Eric Hoffmann, Jonathan Lai, Raynald Chung, Kai Yang, Nihal Balani, Arthur Bražinskas, Andrei Sozanschi, Matthew Hayes, Héctor~Fernández Alcalde, Peter Makarov, Will Chen, Antonio Stella, Liselotte Snijders, Michael Mandl, Ante Kärrman, Paweł Nowak, Xinyi Wu, Alex Dyck, Krishnan Vaidyanathan, Raghavender R, Jessica Mallet, Mitch Rudominer, Eric Johnston, Sushil Mittal, Akhil Udathu, Janara Christensen, Vishal Verma, Zach Irving, Andreas Santucci, Gamaleldin Elsayed, Elnaz Davoodi, Marin Georgiev, Ian Tenney, Nan Hua, Geoffrey Cideron, Edouard Leurent, Mahmoud Alnahlawi, Ionut Georgescu, Nan Wei, Ivy Zheng, Dylan Scandinaro, Heinrich Jiang, Jasper Snoek, Mukund Sundararajan, Xuezhi Wang, Zack Ontiveros, Itay Karo, Jeremy Cole, Vinu Rajashekhar, Lara Tumeh, Eyal
  Ben-David, Rishub Jain, Jonathan Uesato, Romina Datta, Oskar Bunyan, Shimu Wu, John Zhang, Piotr Stanczyk, Ye~Zhang, David Steiner, Subhajit Naskar, Michael Azzam, Matthew Johnson, Adam Paszke, Chung-Cheng Chiu, Jaume~Sanchez Elias, Afroz Mohiuddin, Faizan Muhammad, Jin Miao, Andrew Lee, Nino Vieillard, Jane Park, Jiageng Zhang, Jeff Stanway, Drew Garmon, Abhijit Karmarkar, Zhe Dong, Jong Lee, Aviral Kumar, Luowei Zhou, Jonathan Evens, William Isaac, Geoffrey Irving, Edward Loper, Michael Fink, Isha Arkatkar, Nanxin Chen, Izhak Shafran, Ivan Petrychenko, Zhe Chen, Johnson Jia, Anselm Levskaya, Zhenkai Zhu, Peter Grabowski, Yu~Mao, Alberto Magni, Kaisheng Yao, Javier Snaider, Norman Casagrande, Evan Palmer, Paul Suganthan, Alfonso Castaño, Irene Giannoumis, Wooyeol Kim, Mikołaj Rybiński, Ashwin Sreevatsa, Jennifer Prendki, David Soergel, Adrian Goedeckemeyer, Willi Gierke, Mohsen Jafari, Meenu Gaba, Jeremy Wiesner, Diana~Gage Wright, Yawen Wei, Harsha Vashisht, Yana Kulizhskaya, Jay Hoover, Maigo Le,
  Lu~Li, Chimezie Iwuanyanwu, Lu~Liu, Kevin Ramirez, Andrey Khorlin, Albert Cui, Tian LIN, Marcus Wu, Ricardo Aguilar, Keith Pallo, Abhishek Chakladar, Ginger Perng, Elena~Allica Abellan, Mingyang Zhang, Ishita Dasgupta, Nate Kushman, Ivo Penchev, Alena Repina, Xihui Wu, Tom van~der Weide, Priya Ponnapalli, Caroline Kaplan, Jiri Simsa, Shuangfeng Li, Olivier Dousse, Fan Yang, Jeff Piper, Nathan Ie, Rama Pasumarthi, Nathan Lintz, Anitha Vijayakumar, Daniel Andor, Pedro Valenzuela, Minnie Lui, Cosmin Paduraru, Daiyi Peng, Katherine Lee, Shuyuan Zhang, Somer Greene, Duc~Dung Nguyen, Paula Kurylowicz, Cassidy Hardin, Lucas Dixon, Lili Janzer, Kiam Choo, Ziqiang Feng, Biao Zhang, Achintya Singhal, Dayou Du, Dan McKinnon, Natasha Antropova, Tolga Bolukbasi, Orgad Keller, David Reid, Daniel Finchelstein, Maria~Abi Raad, Remi Crocker, Peter Hawkins, Robert Dadashi, Colin Gaffney, Ken Franko, Anna Bulanova, Rémi Leblond, Shirley Chung, Harry Askham, Luis~C. Cobo, Kelvin Xu, Felix Fischer, Jun Xu, Christina Sorokin,
  Chris Alberti, Chu-Cheng Lin, Colin Evans, Alek Dimitriev, Hannah Forbes, Dylan Banarse, Zora Tung, Mark Omernick, Colton Bishop, Rachel Sterneck, Rohan Jain, Jiawei Xia, Ehsan Amid, Francesco Piccinno, Xingyu Wang, Praseem Banzal, Daniel~J. Mankowitz, Alex Polozov, Victoria Krakovna, Sasha Brown, MohammadHossein Bateni, Dennis Duan, Vlad Firoiu, Meghana Thotakuri, Tom Natan, Matthieu Geist, Ser tan Girgin, Hui Li, Jiayu Ye, Ofir Roval, Reiko Tojo, Michael Kwong, James Lee-Thorp, Christopher Yew, Danila Sinopalnikov, Sabela Ramos, John Mellor, Abhishek Sharma, Kathy Wu, David Miller, Nicolas Sonnerat, Denis Vnukov, Rory Greig, Jennifer Beattie, Emily Caveness, Libin Bai, Julian Eisenschlos, Alex Korchemniy, Tomy Tsai, Mimi Jasarevic, Weize Kong, Phuong Dao, Zeyu Zheng, Frederick Liu, Fan Yang, Rui Zhu, Tian~Huey Teh, Jason Sanmiya, Evgeny Gladchenko, Nejc Trdin, Daniel Toyama, Evan Rosen, Sasan Tavakkol, Linting Xue, Chen Elkind, Oliver Woodman, John Carpenter, George Papamakarios, Rupert Kemp, Sushant
  Kafle, Tanya Grunina, Rishika Sinha, Alice Talbert, Diane Wu, Denese Owusu-Afriyie, Cosmo Du, Chloe Thornton, Jordi Pont-Tuset, Pradyumna Narayana, Jing Li, Saaber Fatehi, John Wieting, Omar Ajmeri, Benigno Uria, Yeongil Ko, Laura Knight, Amélie Héliou, Ning Niu, Shane Gu, Chenxi Pang, Yeqing Li, Nir Levine, Ariel Stolovich, Rebeca Santamaria-Fernandez, Sonam Goenka, Wenny Yustalim, Robin Strudel, Ali Elqursh, Charlie Deck, Hyo Lee, Zonglin Li, Kyle Levin, Raphael Hoffmann, Dan Holtmann-Rice, Olivier Bachem, Sho Arora, Christy Koh, Soheil~Hassas Yeganeh, Siim Põder, Mukarram Tariq, Yanhua Sun, Lucian Ionita, Mojtaba Seyedhosseini, Pouya Tafti, Zhiyu Liu, Anmol Gulati, Jasmine Liu, Xinyu Ye, Bart Chrzaszcz, Lily Wang, Nikhil Sethi, Tianrun Li, Ben Brown, Shreya Singh, Wei Fan, Aaron Parisi, Joe Stanton, Vinod Koverkathu, Christopher~A. Choquette-Choo, Yunjie Li, TJ~Lu, Abe Ittycheriah, Prakash Shroff, Mani Varadarajan, Sanaz Bahargam, Rob Willoughby, David Gaddy, Guillaume Desjardins, Marco Cornero, Brona
  Robenek, Bhavishya Mittal, Ben Albrecht, Ashish Shenoy, Fedor Moiseev, Henrik Jacobsson, Alireza Ghaffarkhah, Morgane Rivière, Alanna Walton, Clément Crepy, Alicia Parrish, Zongwei Zhou, Clement Farabet, Carey Radebaugh, Praveen Srinivasan, Claudia van~der Salm, Andreas Fidjeland, Salvatore Scellato, Eri Latorre-Chimoto, Hanna Klimczak-Plucińska, David Bridson, Dario de~Cesare, Tom Hudson, Piermaria Mendolicchio, Lexi Walker, Alex Morris, Matthew Mauger, Alexey Guseynov, Alison Reid, Seth Odoom, Lucia Loher, Victor Cotruta, Madhavi Yenugula, Dominik Grewe, Anastasia Petrushkina, Tom Duerig, Antonio Sanchez, Steve Yadlowsky, Amy Shen, Amir Globerson, Lynette Webb, Sahil Dua, Dong Li, Surya Bhupatiraju, Dan Hurt, Haroon Qureshi, Ananth Agarwal, Tomer Shani, Matan Eyal, Anuj Khare, Shreyas~Rammohan Belle, Lei Wang, Chetan Tekur, Mihir~Sanjay Kale, Jinliang Wei, Ruoxin Sang, Brennan Saeta, Tyler Liechty, Yi~Sun, Yao Zhao, Stephan Lee, Pandu Nayak, Doug Fritz, Manish~Reddy Vuyyuru, John Aslanides, Nidhi Vyas,
  Martin Wicke, Xiao Ma, Evgenii Eltyshev, Nina Martin, Hardie Cate, James Manyika, Keyvan Amiri, Yelin Kim, Xi~Xiong, Kai Kang, Florian Luisier, Nilesh Tripuraneni, David Madras, Mandy Guo, Austin Waters, Oliver Wang, Joshua Ainslie, Jason Baldridge, Han Zhang, Garima Pruthi, Jakob Bauer, Feng Yang, Riham Mansour, Jason Gelman, Yang Xu, George Polovets, Ji~Liu, Honglong Cai, Warren Chen, XiangHai Sheng, Emily Xue, Sherjil Ozair, Christof Angermueller, Xiaowei Li, Anoop Sinha, Weiren Wang, Julia Wiesinger, Emmanouil Koukoumidis, Yuan Tian, Anand Iyer, Madhu Gurumurthy, Mark Goldenson, Parashar Shah, MK~Blake, Hongkun Yu, Anthony Urbanowicz, Jennimaria Palomaki, Chrisantha Fernando, Ken Durden, Harsh Mehta, Nikola Momchev, Elahe Rahimtoroghi, Maria Georgaki, Amit Raul, Sebastian Ruder, Morgan Redshaw, Jinhyuk Lee, Denny Zhou, Komal Jalan, Dinghua Li, Blake Hechtman, Parker Schuh, Milad Nasr, Kieran Milan, Vladimir Mikulik, Juliana Franco, Tim Green, Nam Nguyen, Joe Kelley, Aroma Mahendru, Andrea Hu, Joshua
  Howland, Ben Vargas, Jeffrey Hui, Kshitij Bansal, Vikram Rao, Rakesh Ghiya, Emma Wang, Ke~Ye, Jean~Michel Sarr, Melanie~Moranski Preston, Madeleine Elish, Steve Li, Aakash Kaku, Jigar Gupta, Ice Pasupat, Da-Cheng Juan, Milan Someswar, Tejvi M., Xinyun Chen, Aida Amini, Alex Fabrikant, Eric Chu, Xuanyi Dong, Amruta Muthal, Senaka Buthpitiya, Sarthak Jauhari, Nan Hua, Urvashi Khandelwal, Ayal Hitron, Jie Ren, Larissa Rinaldi, Shahar Drath, Avigail Dabush, Nan-Jiang Jiang, Harshal Godhia, Uli Sachs, Anthony Chen, Yicheng Fan, Hagai Taitelbaum, Hila Noga, Zhuyun Dai, James Wang, Chen Liang, Jenny Hamer, Chun-Sung Ferng, Chenel Elkind, Aviel Atias, Paulina Lee, Vít Listík, Mathias Carlen, Jan van~de Kerkhof, Marcin Pikus, Krunoslav Zaher, Paul Müller, Sasha Zykova, Richard Stefanec, Vitaly Gatsko, Christoph Hirnschall, Ashwin Sethi, Xingyu~Federico Xu, Chetan Ahuja, Beth Tsai, Anca Stefanoiu, Bo~Feng, Keshav Dhandhania, Manish Katyal, Akshay Gupta, Atharva Parulekar, Divya Pitta, Jing Zhao, Vivaan Bhatia,
  Yashodha Bhavnani, Omar Alhadlaq, Xiaolin Li, Peter Danenberg, Dennis Tu, Alex Pine, Vera Filippova, Abhipso Ghosh, Ben Limonchik, Bhargava Urala, Chaitanya~Krishna Lanka, Derik Clive, Yi~Sun, Edward Li, Hao Wu, Kevin Hongtongsak, Ianna Li, Kalind Thakkar, Kuanysh Omarov, Kushal Majmundar, Michael Alverson, Michael Kucharski, Mohak Patel, Mudit Jain, Maksim Zabelin, Paolo Pelagatti, Rohan Kohli, Saurabh Kumar, Joseph Kim, Swetha Sankar, Vineet Shah, Lakshmi Ramachandruni, Xiangkai Zeng, Ben Bariach, Laura Weidinger, Amar Subramanya, Sissie Hsiao, Demis Hassabis, Koray Kavukcuoglu, Adam Sadovsky, Quoc Le, Trevor Strohman, Yonghui Wu, Slav Petrov, Jeffrey Dean, and Oriol Vinyals.
\newblock Gemini: A family of highly capable multimodal models, 2024.

\bibitem[Gemini~Team et~al.(2024)Gemini~Team, Mesnard, Hardin, Dadashi, Bhupatiraju, Pathak, Sifre, Rivi{\`e}re, Kale, Love, et~al.]{team2024gemma}
Gemma Gemini~Team, Thomas Mesnard, Cassidy Hardin, Robert Dadashi, Surya Bhupatiraju, Shreya Pathak, Laurent Sifre, Morgane Rivi{\`e}re, Mihir~Sanjay Kale, Juliette Love, et~al.
\newblock Gemma: Open models based on gemini research and technology.
\newblock \emph{arXiv preprint arXiv:2403.08295}, 2024.

\bibitem[Gemma-Team(2024)]{gemmareport}
Gemma-Team.
\newblock Gemma: Open models based on gemini research and technology, 2024.

\bibitem[Goyal et~al.(2021)Goyal, Gao, Chaudhary, Chen, Wenzek, Ju, Krishnan, Ranzato, Guzman, and Fan]{goyal2021flores101}
Naman Goyal, Cynthia Gao, Vishrav Chaudhary, Peng-Jen Chen, Guillaume Wenzek, Da~Ju, Sanjana Krishnan, Marc'Aurelio Ranzato, Francisco Guzman, and Angela Fan.
\newblock The flores-101 evaluation benchmark for low-resource and multilingual machine translation.
\newblock \emph{arXiv}, abs/2106.03193, 2021.

\bibitem[Hasan et~al.(2021)Hasan, Bhattacharjee, Islam, Samin, Li, Kang, Rahman, and Shahriyar]{2021_hasanXLSumLargeScaleMultilingual}
Tahmid Hasan, Abhik Bhattacharjee, Md~Saiful Islam, Kazi Samin, Yuan-Fang Li, Yong-Bin Kang, M.~Sohel Rahman, and Rifat Shahriyar.
\newblock {{XL-Sum}}: {{Large-Scale Multilingual Abstractive Summarization}} for 44 {{Languages}}.
\newblock pp.\  4693--4703, August 2021.
\newblock \doi{10.48550/arXiv.2106.13822}.
\newblock URL \url{https://aclanthology.org/2021.findings-acl.413}.

\bibitem[Held et~al.(2023)Held, Harris, Best, and Yang]{held2023material}
William Held, Camille Harris, Michael Best, and Diyi Yang.
\newblock A material lens on coloniality in nlp.
\newblock \emph{arXiv}, abs/2311.08391, 2023.

\bibitem[Hendrycks et~al.(2020)Hendrycks, Burns, Basart, Zou, Mazeika, Song, and Steinhardt]{hendrycks2020measuring}
Dan Hendrycks, Collin Burns, Steven Basart, Andy Zou, Mantas Mazeika, Dawn Song, and Jacob Steinhardt.
\newblock Measuring massive multitask language understanding.
\newblock In \emph{International Conference on Learning Representations}, 2020.

\bibitem[Jiang et~al.(2023)Jiang, Sablayrolles, Mensch, Bamford, Chaplot, de~las Casas, Bressand, Lengyel, Lample, Saulnier, Lavaud, Lachaux, Stock, Scao, Lavril, Wang, Lacroix, and Sayed]{jiang2023mistral}
Albert~Q. Jiang, Alexandre Sablayrolles, Arthur Mensch, Chris Bamford, Devendra~Singh Chaplot, Diego de~las Casas, Florian Bressand, Gianna Lengyel, Guillaume Lample, Lucile Saulnier, Lélio~Renard Lavaud, Marie-Anne Lachaux, Pierre Stock, Teven~Le Scao, Thibaut Lavril, Thomas Wang, Timothée Lacroix, and William~El Sayed.
\newblock Mistral 7b, 2023.

\bibitem[Jiang et~al.(2024)Jiang, Sablayrolles, Roux, Mensch, Savary, Bamford, Chaplot, de~las Casas, Hanna, Bressand, Lengyel, Bour, Lample, Lavaud, Saulnier, Lachaux, Stock, Subramanian, Yang, Antoniak, Scao, Gervet, Lavril, Wang, Lacroix, and Sayed]{jiang2024mixtral}
Albert~Q. Jiang, Alexandre Sablayrolles, Antoine Roux, Arthur Mensch, Blanche Savary, Chris Bamford, Devendra~Singh Chaplot, Diego de~las Casas, Emma~Bou Hanna, Florian Bressand, Gianna Lengyel, Guillaume Bour, Guillaume Lample, Lélio~Renard Lavaud, Lucile Saulnier, Marie-Anne Lachaux, Pierre Stock, Sandeep Subramanian, Sophia Yang, Szymon Antoniak, Teven~Le Scao, Théophile Gervet, Thibaut Lavril, Thomas Wang, Timothée Lacroix, and William~El Sayed.
\newblock Mixtral of experts.
\newblock \emph{arXiv}, abs/2401.04088, 2024.

\bibitem[Jouppi et~al.(2023)Jouppi, Kurian, Li, Ma, Nagarajan, Nai, Patil, Subramanian, Swing, Towles, Young, Zhou, Zhou, and Patterson]{tpuv4}
Norman~P. Jouppi, George Kurian, Sheng Li, Peter Ma, Rahul Nagarajan, Lifeng Nai, Nishant Patil, Suvinay Subramanian, Andy Swing, Brian Towles, Cliff Young, Xiang Zhou, Zongwei Zhou, and David Patterson.
\newblock Tpu v4: An optically reconfigurable supercomputer for machine learning with hardware support for embeddings, 2023.

\bibitem[Khandelwal et~al.(2023)Khandelwal, Tonneau, Bean, Kirk, and Hale]{Khandelwal2023CasteistBN}
Khyati Khandelwal, Manuel Tonneau, Andrew~M. Bean, Hannah~Rose Kirk, and Scott~A. Hale.
\newblock Casteist but not racist? quantifying disparities in large language model bias between india and the west.
\newblock \emph{ArXiv}, abs/2309.08573, 2023.
\newblock URL \url{https://api.semanticscholar.org/CorpusID:262013517}.

\bibitem[Khondaker et~al.(2023)Khondaker, Waheed, Nagoudi, and Abdul-Mageed]{khondaker2023gptaraeval}
Md~Tawkat~Islam Khondaker, Abdul Waheed, El~Moatez~Billah Nagoudi, and Muhammad Abdul-Mageed.
\newblock Gptaraeval: A comprehensive evaluation of chatgpt on arabic nlp.
\newblock \emph{arXiv}, abs/2305.14976, 2023.

\bibitem[Kim et~al.(2023)Kim, Shin, Cho, Jang, Longpre, Lee, Yun, Shin, Kim, Thorne, et~al.]{kim2023prometheus}
Seungone Kim, Jamin Shin, Yejin Cho, Joel Jang, Shayne Longpre, Hwaran Lee, Sangdoo Yun, Seongjin Shin, Sungdong Kim, James Thorne, et~al.
\newblock Prometheus: Inducing fine-grained evaluation capability in language models.
\newblock \emph{arXiv preprint arXiv:2310.08491}, 2023.

\bibitem[Kingma \& Ba(2014)Kingma and Ba]{kingma2014adam}
Diederik~P Kingma and Jimmy Ba.
\newblock {Adam: A method for stochastic optimization}.
\newblock \emph{arXiv preprint arXiv:1412.6980}, 2014.

\bibitem[Kotek et~al.(2023)Kotek, Dockum, and Sun]{Kotek2023GenderBA}
Hadas Kotek, Rikker Dockum, and David~Q. Sun.
\newblock Gender bias and stereotypes in large language models.
\newblock \emph{Proceedings of The ACM Collective Intelligence Conference}, 2023.
\newblock URL \url{https://api.semanticscholar.org/CorpusID:261276445}.

\bibitem[Li et~al.(2023{\natexlab{a}})Li, Koto, Wu, Aji, and Baldwin]{li2023bactrianx}
Haonan Li, Fajri Koto, Minghao Wu, Alham~Fikri Aji, and Timothy Baldwin.
\newblock Bactrian-x: Multilingual replicable instruction-following models with low-rank adaptation.
\newblock \emph{arXiv}, abs/2305.15011, 2023{\natexlab{a}}.

\bibitem[Li et~al.(2023{\natexlab{b}})Li, Chen, Luo, Kang, Zhang, Hu, Chan, and Song]{Li2023PrivacyIL}
Haoran Li, Yulin Chen, Jinglong Luo, Yan Kang, Xiaojin Zhang, Qi~Hu, Chunkit Chan, and Yangqiu Song.
\newblock Privacy in large language models: Attacks, defenses and future directions.
\newblock \emph{ArXiv}, abs/2310.10383, 2023{\natexlab{b}}.
\newblock URL \url{https://api.semanticscholar.org/CorpusID:264145758}.

\bibitem[Lin et~al.(2021)Lin, Mihaylov, Artetxe, Wang, Chen, Simig, Ott, Goyal, Bhosale, Du, Pasunuru, Shleifer, Koura, Chaudhary, O'Horo, Wang, Zettlemoyer, Kozareva, Diab, Stoyanov, and Li]{lin2021fewshot}
Xi~Victoria Lin, Todor Mihaylov, Mikel Artetxe, Tianlu Wang, Shuohui Chen, Daniel Simig, Myle Ott, Naman Goyal, Shruti Bhosale, Jingfei Du, Ramakanth Pasunuru, Sam Shleifer, Punit~Singh Koura, Vishrav Chaudhary, Brian O'Horo, Jeff Wang, Luke Zettlemoyer, Zornitsa Kozareva, Mona Diab, Veselin Stoyanov, and Xian Li.
\newblock Few-shot learning with multilingual language models.
\newblock \emph{arXiv}, abs/2112.10668, 2021.

\bibitem[Longpre et~al.(2023{\natexlab{a}})Longpre, Hou, Vu, Webson, Chung, Tay, Zhou, Le, Zoph, Wei, and Roberts]{longpre2023flan}
Shayne Longpre, Le~Hou, Tu~Vu, Albert Webson, Hyung~Won Chung, Yi~Tay, Denny Zhou, Quoc~V. Le, Barret Zoph, Jason Wei, and Adam Roberts.
\newblock The flan collection: Designing data and methods for effective instruction tuning.
\newblock \emph{arXiv}, abs/2301.13688, 2023{\natexlab{a}}.

\bibitem[Longpre et~al.(2023{\natexlab{b}})Longpre, Mahari, Chen, Obeng-Marnu, Sileo, Brannon, Muennighoff, Khazam, Kabbara, Perisetla, et~al.]{longpre2023data}
Shayne Longpre, Robert Mahari, Anthony Chen, Naana Obeng-Marnu, Damien Sileo, William Brannon, Niklas Muennighoff, Nathan Khazam, Jad Kabbara, Kartik Perisetla, et~al.
\newblock The data provenance initiative: A large scale audit of dataset licensing \& attribution in ai.
\newblock \emph{arXiv preprint arXiv:2310.16787}, 2023{\natexlab{b}}.

\bibitem[Lukas et~al.(2023)Lukas, Salem, Sim, Tople, Wutschitz, and Zanella-B'eguelin]{Lukas2023AnalyzingLO}
Nils Lukas, A.~Salem, Robert Sim, Shruti Tople, Lukas Wutschitz, and Santiago Zanella-B'eguelin.
\newblock Analyzing leakage of personally identifiable information in language models.
\newblock \emph{2023 IEEE Symposium on Security and Privacy (SP)}, pp.\  346--363, 2023.
\newblock URL \url{https://api.semanticscholar.org/CorpusID:256459554}.

\bibitem[Muennighoff et~al.(2023)Muennighoff, Wang, Sutawika, Roberts, Biderman, Le~Scao, Bari, Shen, Yong, Schoelkopf, Tang, Radev, Aji, Almubarak, Albanie, Alyafeai, Webson, Raff, and Raffel]{muennighoff2022crosslingual}
Niklas Muennighoff, Thomas Wang, Lintang Sutawika, Adam Roberts, Stella Biderman, Teven Le~Scao, M~Saiful Bari, Sheng Shen, Zheng~Xin Yong, Hailey Schoelkopf, Xiangru Tang, Dragomir Radev, Alham~Fikri Aji, Khalid Almubarak, Samuel Albanie, Zaid Alyafeai, Albert Webson, Edward Raff, and Colin Raffel.
\newblock Crosslingual generalization through multitask finetuning.
\newblock In Anna Rogers, Jordan Boyd-Graber, and Naoaki Okazaki (eds.), \emph{Proceedings of the 61st Annual Meeting of the Association for Computational Linguistics (Volume 1: Long Papers)}, pp.\  15991--16111, Toronto, Canada, July 2023. Association for Computational Linguistics.
\newblock \doi{10.18653/v1/2023.acl-long.891}.
\newblock URL \url{https://aclanthology.org/2023.acl-long.891}.

\bibitem[Nasr et~al.(2023)Nasr, Carlini, Hayase, Jagielski, Cooper, Ippolito, Choquette-Choo, Wallace, Tramèr, and Lee]{nasr2023scalable}
Milad Nasr, Nicholas Carlini, Jonathan Hayase, Matthew Jagielski, A.~Feder Cooper, Daphne Ippolito, Christopher~A. Choquette-Choo, Eric Wallace, Florian Tramèr, and Katherine Lee.
\newblock Scalable extraction of training data from (production) language models.
\newblock \emph{arXiv}, abs/2311.17035, 2023.

\bibitem[Nicholas \& Bhatia(2023)Nicholas and Bhatia]{nicholas2023lost}
Gabriel Nicholas and Aliya Bhatia.
\newblock Lost in translation: Large language models in non-english content analysis.
\newblock \emph{arXiv}, abs/2306.07377, 2023.

\bibitem[NLLB-Team et~al.(2022)NLLB-Team, Costa-jussà, Cross, Çelebi, Elbayad, Heafield, Heffernan, Kalbassi, Lam, Licht, Maillard, Sun, Wang, Wenzek, Youngblood, Akula, Barrault, Gonzalez, Hansanti, Hoffman, Jarrett, Sadagopan, Rowe, Spruit, Tran, Andrews, Ayan, Bhosale, Edunov, Fan, Gao, Goswami, Guzmán, Koehn, Mourachko, Ropers, Saleem, Schwenk, and Wang]{nllb2022}
NLLB-Team, Marta~R. Costa-jussà, James Cross, Onur Çelebi, Maha Elbayad, Kenneth Heafield, Kevin Heffernan, Elahe Kalbassi, Janice Lam, Daniel Licht, Jean Maillard, Anna Sun, Skyler Wang, Guillaume Wenzek, Al~Youngblood, Bapi Akula, Loic Barrault, Gabriel~Mejia Gonzalez, Prangthip Hansanti, John Hoffman, Semarley Jarrett, Kaushik~Ram Sadagopan, Dirk Rowe, Shannon Spruit, Chau Tran, Pierre Andrews, Necip~Fazil Ayan, Shruti Bhosale, Sergey Edunov, Angela Fan, Cynthia Gao, Vedanuj Goswami, Francisco Guzmán, Philipp Koehn, Alexandre Mourachko, Christophe Ropers, Safiyyah Saleem, Holger Schwenk, and Jeff Wang.
\newblock No language left behind: Scaling human-centered machine translation.
\newblock 2022.

\bibitem[Ojo et~al.(2023)Ojo, Ogueji, Stenetorp, and Adelani]{ojo2023good}
Jessica Ojo, Kelechi Ogueji, Pontus Stenetorp, and David~I. Adelani.
\newblock How good are large language models on african languages?
\newblock \emph{arXiv}, abs/2311.07978, 2023.

\bibitem[Pfeiffer et~al.(2022)Pfeiffer, Goyal, Lin, Li, Cross, Riedel, and Artetxe]{pfeiffer2022lifting}
Jonas Pfeiffer, Naman Goyal, Xi~Lin, Xian Li, James Cross, Sebastian Riedel, and Mikel Artetxe.
\newblock Lifting the curse of multilinguality by pre-training modular transformers.
\newblock In \emph{Proceedings of the 2022 Conference of the North American Chapter of the Association for Computational Linguistics: Human Language Technologies}, pp.\  3479--3495, Seattle, United States, July 2022. Association for Computational Linguistics.
\newblock \doi{10.18653/v1/2022.naacl-main.255}.
\newblock URL \url{https://aclanthology.org/2022.naacl-main.255}.

\bibitem[Ponti et~al.(2020)Ponti, Glava{\v{s}}, Majewska, Liu, Vuli{\'c}, and Korhonen]{ponti2020xcopa}
Edoardo~Maria Ponti, Goran Glava{\v{s}}, Olga Majewska, Qianchu Liu, Ivan Vuli{\'c}, and Anna Korhonen.
\newblock Xcopa: A multilingual dataset for causal commonsense reasoning.
\newblock pp.\  2362--2376, November 2020.
\newblock \doi{10.18653/v1/2020.emnlp-main.185}.
\newblock URL \url{https://aclanthology.org/2020.emnlp-main.185}.

\bibitem[Press et~al.(2021)Press, Smith, and Lewis]{alibi}
Ofir Press, Noah~A. Smith, and Mike Lewis.
\newblock Train short, test long: Attention with linear biases enables input length extrapolation.
\newblock \emph{CoRR}, abs/2108.12409, 2021.
\newblock URL \url{https://arxiv.org/abs/2108.12409}.

\bibitem[Rafailov et~al.(2023)Rafailov, Sharma, Mitchell, Ermon, Manning, and Finn]{rafailov2023direct}
Rafael Rafailov, Archit Sharma, Eric Mitchell, Stefano Ermon, Christopher~D Manning, and Chelsea Finn.
\newblock Direct preference optimization: Your language model is secretly a reward model.
\newblock \emph{arXiv preprint arXiv:2305.18290}, 2023.

\bibitem[Schwartz et~al.(2022)Schwartz, Vassilev, Greene, Perine, Burt, Hall, et~al.]{schwartz2022towards}
Reva Schwartz, Apostol Vassilev, Kristen Greene, Lori Perine, Andrew Burt, Patrick Hall, et~al.
\newblock Towards a standard for identifying and managing bias in artificial intelligence.
\newblock \emph{NIST special publication}, 1270\penalty0 (10.6028), 2022.

\bibitem[Shazeer(2020)]{gluvariants}
Noam Shazeer.
\newblock {GLU} variants improve transformer.
\newblock \emph{CoRR}, abs/2002.05202, 2020.
\newblock URL \url{https://arxiv.org/abs/2002.05202}.

\bibitem[Shi et~al.(2023)Shi, Suzgun, Freitag, Wang, Srivats, Vosoughi, Chung, Tay, Ruder, Zhou, Das, and Wei]{shi2023language-mgsm}
Freda Shi, Mirac Suzgun, Markus Freitag, Xuezhi Wang, Suraj Srivats, Soroush Vosoughi, Hyung~Won Chung, Yi~Tay, Sebastian Ruder, Denny Zhou, Dipanjan Das, and Jason Wei.
\newblock Language models are multilingual chain-of-thought reasoners.
\newblock In \emph{The Eleventh International Conference on Learning Representations}, 2023.
\newblock URL \url{https://openreview.net/forum?id=fR3wGCk-IXp}.

\bibitem[Singh et~al.(2024)Singh, Vargus, Dsouza, Karlsson, Mahendiran, Ko, Shandilya, Patel, Mataciunas, OMahony, Zhang, Hettiarachchi, Wilson, Machado, Moura, Krzemiński, Fadaei, Ergün, Okoh, Alaagib, Mudannayake, Alyafeai, Chien, Ruder, Guthikonda, Alghamdi, Gehrmann, Muennighoff, Bartolo, Kreutzer, Üstün, Fadaee, and Hooker]{ayadata2024}
Shivalika Singh, Freddie Vargus, Daniel Dsouza, Börje~F. Karlsson, Abinaya Mahendiran, Wei-Yin Ko, Herumb Shandilya, Jay Patel, Deividas Mataciunas, Laura OMahony, Mike Zhang, Ramith Hettiarachchi, Joseph Wilson, Marina Machado, Luisa~Souza Moura, Dominik Krzemiński, Hakimeh Fadaei, Irem Ergün, Ifeoma Okoh, Aisha Alaagib, Oshan Mudannayake, Zaid Alyafeai, Vu~Minh Chien, Sebastian Ruder, Surya Guthikonda, Emad~A. Alghamdi, Sebastian Gehrmann, Niklas Muennighoff, Max Bartolo, Julia Kreutzer, Ahmet Üstün, Marzieh Fadaee, and Sara Hooker.
\newblock Aya dataset: An open-access collection for multilingual instruction tuning.
\newblock \emph{arXiv preprint arXiv:2402.06619}, 2024.

\bibitem[Su et~al.(2021)Su, Lu, Pan, Wen, and Liu]{rope}
Jianlin Su, Yu~Lu, Shengfeng Pan, Bo~Wen, and Yunfeng Liu.
\newblock Roformer: Enhanced transformer with rotary position embedding.
\newblock \emph{CoRR}, abs/2104.09864, 2021.
\newblock URL \url{https://arxiv.org/abs/2104.09864}.

\bibitem[Taori et~al.(2023)Taori, Gulrajani, Zhang, Dubois, Li, Guestrin, Liang, and Hashimoto]{taori2023stanford}
Rohan Taori, Ishaan Gulrajani, Tianyi Zhang, Yann Dubois, Xuechen Li, Carlos Guestrin, Percy Liang, and Tatsunori~B Hashimoto.
\newblock Stanford alpaca: An instruction-following llama model.
\newblock 2023.

\bibitem[Touvron et~al.(2023{\natexlab{a}})Touvron, Lavril, Izacard, Martinet, Lachaux, Lacroix, Rozière, Goyal, Hambro, Azhar, Rodriguez, Joulin, Grave, and Lample]{touvron2023llama}
Hugo Touvron, Thibaut Lavril, Gautier Izacard, Xavier Martinet, Marie-Anne Lachaux, Timothée Lacroix, Baptiste Rozière, Naman Goyal, Eric Hambro, Faisal Azhar, Aurelien Rodriguez, Armand Joulin, Edouard Grave, and Guillaume Lample.
\newblock Llama: Open and efficient foundation language models.
\newblock \emph{arXiv}, abs/2302.13971, 2023{\natexlab{a}}.

\bibitem[Touvron et~al.(2023{\natexlab{b}})Touvron, Martin, Stone, Albert, Almahairi, Babaei, Bashlykov, Batra, Bhargava, Bhosale, Bikel, Blecher, Ferrer, Chen, Cucurull, Esiobu, Fernandes, Fu, Fu, Fuller, Gao, Goswami, Goyal, Hartshorn, Hosseini, Hou, Inan, Kardas, Kerkez, Khabsa, Kloumann, Korenev, Koura, Lachaux, Lavril, Lee, Liskovich, Lu, Mao, Martinet, Mihaylov, Mishra, Molybog, Nie, Poulton, Reizenstein, Rungta, Saladi, Schelten, Silva, Smith, Subramanian, Tan, Tang, Taylor, Williams, Kuan, Xu, Yan, Zarov, Zhang, Fan, Kambadur, Narang, Rodriguez, Stojnic, Edunov, and Scialom]{touvron2023llama2}
Hugo Touvron, Louis Martin, Kevin Stone, Peter Albert, Amjad Almahairi, Yasmine Babaei, Nikolay Bashlykov, Soumya Batra, Prajjwal Bhargava, Shruti Bhosale, Dan Bikel, Lukas Blecher, Cristian~Canton Ferrer, Moya Chen, Guillem Cucurull, David Esiobu, Jude Fernandes, Jeremy Fu, Wenyin Fu, Brian Fuller, Cynthia Gao, Vedanuj Goswami, Naman Goyal, Anthony Hartshorn, Saghar Hosseini, Rui Hou, Hakan Inan, Marcin Kardas, Viktor Kerkez, Madian Khabsa, Isabel Kloumann, Artem Korenev, Punit~Singh Koura, Marie-Anne Lachaux, Thibaut Lavril, Jenya Lee, Diana Liskovich, Yinghai Lu, Yuning Mao, Xavier Martinet, Todor Mihaylov, Pushkar Mishra, Igor Molybog, Yixin Nie, Andrew Poulton, Jeremy Reizenstein, Rashi Rungta, Kalyan Saladi, Alan Schelten, Ruan Silva, Eric~Michael Smith, Ranjan Subramanian, Xiaoqing~Ellen Tan, Binh Tang, Ross Taylor, Adina Williams, Jian~Xiang Kuan, Puxin Xu, Zheng Yan, Iliyan Zarov, Yuchen Zhang, Angela Fan, Melanie Kambadur, Sharan Narang, Aurelien Rodriguez, Robert Stojnic, Sergey Edunov, and Thomas
  Scialom.
\newblock Llama 2: Open foundation and fine-tuned chat models.
\newblock \emph{arXiv}, abs/2307.09288, 2023{\natexlab{b}}.

\bibitem[Vashishtha et~al.(2023)Vashishtha, Ahuja, and Sitaram]{vashishtha2023evaluating}
Aniket Vashishtha, Kabir Ahuja, and Sunayana Sitaram.
\newblock On evaluating and mitigating gender biases in multilingual settings.
\newblock \emph{arXiv}, abs/2307.01503, 2023.

\bibitem[Xue et~al.(2020)Xue, Constant, Roberts, Kale, Al-Rfou, Siddhant, Barua, and Raffel]{xue2020mt5}
Linting Xue, Noah Constant, Adam Roberts, Mihir Kale, Rami Al-Rfou, Aditya Siddhant, Aditya Barua, and Colin Raffel.
\newblock mt5: A massively multilingual pre-trained text-to-text transformer.
\newblock pp.\  483--498, June 2020.
\newblock \doi{10.18653/v1/2021.naacl-main.41}.
\newblock URL \url{https://aclanthology.org/2021.naacl-main.41}.

\bibitem[Yang et~al.(2018)Yang, Qi, Zhang, Bengio, Cohen, Salakhutdinov, and Manning]{yang2018hotpotqa}
Zhilin Yang, Peng Qi, Saizheng Zhang, Yoshua Bengio, William~W. Cohen, Ruslan Salakhutdinov, and Christopher~D. Manning.
\newblock {HotpotQA}: A dataset for diverse, explainable multi-hop question answering.
\newblock In \emph{Conference on Empirical Methods in Natural Language Processing ({EMNLP})}, pp.\  2369--2380, Brussels, Belgium, October-November 2018. Association for Computational Linguistics.
\newblock \doi{10.18653/v1/D18-1259}.
\newblock URL \url{https://aclanthology.org/D18-1259}.

\bibitem[Yong et~al.(2023{\natexlab{a}})Yong, Menghini, and Bach]{yong2023lowresource}
Zheng-Xin Yong, Cristina Menghini, and Stephen~H. Bach.
\newblock Low-resource languages jailbreak {GPT}-4.
\newblock \emph{arXiv}, abs/2310.02446, 2023{\natexlab{a}}.

\bibitem[Yong et~al.(2023{\natexlab{b}})Yong, Schoelkopf, Muennighoff, Aji, Adelani, Almubarak, Bari, Sutawika, Kasai, Baruwa, Winata, Biderman, Raff, Radev, and Nikoulina]{yong2022bloom+}
Zheng~Xin Yong, Hailey Schoelkopf, Niklas Muennighoff, Alham~Fikri Aji, David~Ifeoluwa Adelani, Khalid Almubarak, M~Saiful Bari, Lintang Sutawika, Jungo Kasai, Ahmed Baruwa, Genta Winata, Stella Biderman, Edward Raff, Dragomir Radev, and Vassilina Nikoulina.
\newblock {BLOOM}+1: Adding language support to {BLOOM} for zero-shot prompting.
\newblock In \emph{Proceedings of the 61st Annual Meeting of the Association for Computational Linguistics (Volume 1: Long Papers)}, pp.\  11682--11703, Toronto, Canada, July 2023{\natexlab{b}}. Association for Computational Linguistics.
\newblock \doi{10.18653/v1/2023.acl-long.653}.
\newblock URL \url{https://aclanthology.org/2023.acl-long.653}.

\bibitem[Yoo et~al.(2022)Yoo, Perlin, Kamalakara, and Araújo]{fax}
Joanna Yoo, Kuba Perlin, Siddhartha~Rao Kamalakara, and João G.~M. Araújo.
\newblock Scalable training of language models using jax pjit and tpuv4, 2022.

\bibitem[Zhao et~al.(2024)Zhao, Zhang, Gao, Zhang, Gui, and Huang]{zhao2024llama}
Jun Zhao, Zhihao Zhang, Luhui Gao, Qi~Zhang, Tao Gui, and Xuanjing Huang.
\newblock Llama beyond english: An empirical study on language capability transfer.
\newblock \emph{arXiv}, abs/2401.01055, 2024.

\bibitem[Üstün et~al.(2024)Üstün, Aryabumi, Yong, Ko, D'souza, Onilude, Bhandari, Singh, Ooi, Kayid, Vargus, Blunsom, Longpre, Muennighoff, Fadaee, Kreutzer, and Hooker]{ustun2024aya}
Ahmet Üstün, Viraat Aryabumi, Zheng-Xin Yong, Wei-Yin Ko, Daniel D'souza, Gbemileke Onilude, Neel Bhandari, Shivalika Singh, Hui-Lee Ooi, Amr Kayid, Freddie Vargus, Phil Blunsom, Shayne Longpre, Niklas Muennighoff, Marzieh Fadaee, Julia Kreutzer, and Sara Hooker.
\newblock Aya model: An instruction finetuned open-access multilingual language model, 2024.

\end{thebibliography}
